\documentclass[10pt,twocolumn,letterpaper]{article}
\usepackage[table]{xcolor}
\usepackage{iccv}

\usepackage{times}
\usepackage{epsfig}
\usepackage{graphicx}
\usepackage{amsmath}
\usepackage{amssymb}
\usepackage{url}
\usepackage{epigraph}
\usepackage{subfigure}
\usepackage[pagebackref=true,breaklinks=true,letterpaper=true,colorlinks,bookmarks=false]{hyperref}

\iccvfinalcopy 

\def\iccvPaperID{911} 

\ificcvfinal\fi

\usepackage{etoolbox}
\makeatletter
\patchcmd{\epigraph}{\@epitext{#1}}{\itshape\@epitext{#1}}{}{}
\makeatother

\def\etal{{\em et al.\/}\,}
\begin{document}

\title{Understanding and Diagnosing Visual Tracking Systems}
\author{{Naiyan Wang}$^\dagger$ \quad {Jianping Shi}$^\ddagger$ \quad {Dit-Yan Yeung}$^\dagger$ \quad {Jiaya Jia}$^\ddagger$ \\
	$^\dagger$ Hong Kong University of Science and Technology \quad \quad 
	$^\ddagger$ Chinese University of Hongkong \\
	{\tt\small winsty@gmail.com} \quad {\tt\small jpshi@cse.cuhk.edu.hk}  \quad {\tt\small dyyeung@cse.ust.hk} \quad {\tt \small leojia@cse.cuhk.edu.hk}
} 

\maketitle


\begin{abstract}
Several benchmark datasets for visual tracking research have been proposed in recent years.  Despite their usefulness, whether they are sufficient for understanding and diagnosing the strengths and weaknesses of different trackers remains questionable.  To address this issue, we propose a framework by breaking a tracker down into five constituent parts, namely, motion model, feature extractor, observation model, model updater, and ensemble post-processor.  We then conduct ablative experiments on each component to study how it affects the overall result.  Surprisingly, our findings are discrepant with some common beliefs in the visual tracking research community.  We find that the feature extractor plays the most important role in a tracker.  On the other hand, although the observation model is the focus of many studies, we find that it often brings no significant improvement.  Moreover, the motion model and model updater contain many details that could affect the result.  Also, the ensemble post-processor can improve the result substantially when the constituent trackers have high diversity.  Based on our findings, we put together some very elementary building blocks to give a basic tracker which is competitive in performance to the state-of-the-art trackers.  We believe our framework can provide a solid baseline when conducting controlled experiments for visual tracking research.
\end{abstract}

\section{Introduction}
Visual tracking is an essential building block of many advanced applications in the areas such as video surveillance and human-computer interaction.  In this paper, we focus on the most general type of visual tracking problems, namely, short-term single-object model-free tracking~\cite{vot14}.
Numerous such trackers have been proposed over the past few decades, ranging from the simple KLT tracker~\cite{klt1, klt2} in the 1980s to the recent deep learning trackers~\cite{sodlt,cnnt} which are a lot more complex.

Evaluating and comparing trackers has always been a nontrivial task.  For a long time, researchers usually reported tracking results on a small number of videos based on specific model parameters manually tuned for each video.  Since subjective bias~\cite{bias} in the results can be caused by selection of videos, this practice makes it infeasible to give a fair comparison of different trackers.  To address this fairness concern, several relatively large benchmarks~\cite{benchmark,vot14} and evaluation metrics~\cite{metric} have been proposed recently.  With the aid of these benchmarks, we have witnessed substantial advances in recent years.  However, we would like to raise this question: \emph{Is simply evaluating these trackers on the de facto benchmarks sufficient for understanding and diagnosing their strengths and weaknesses?}

We are afraid that the answer to the above question is not affirmative, for the following reason.  Modern trackers are usually complicated systems made up of several separate components.  When a tracker is evaluated as a whole, we cannot gain a detailed understanding of the effectiveness of each component.  For illustration, suppose tracker~A uses histograms of oriented gradients (HOG)~\cite{hog} as features and the support vector machine (SVM) as the observation model, while tracker~B uses raw pixels as features and logistic regression as the observation model.  If tracker~A outperforms tracker~B in a benchmark, can we conclude that SVM is better than logistic regression for tracking?  Obviously drawing such a conclusion would be arbitrary since HOG features have stronger representational power than raw pixels.  This calls for a more carefully designed framework for the evaluation and comparison of trackers.

We propose in this paper a new way to understand and diagnose visual trackers.  Note that our goal is not to create a new benchmark.  Instead, our analysis will still be based on existing benchmarks.  We first break a tracker down into its constituent parts, namely, motion model, feature extractor, observation model, model updater, and ensemble post-processor. 
We note that most existing trackers can be viewed this way.  Based on this framework, we conduct an ablative analysis on a tracker to identify the constituent part that is most crucial to the overall performance of the tracker.  Contrary to popular belief, it turns out that the observation model (which is the focus of many papers on visual tracking) does not play the most important role in a tracker.  Instead, we find that actually the feature extractor affects the performance most.  Moreover, the ensemble post-processor is a simple yet effective way to achieve significant performance boost, but it is comparatively less studied. Furthermore, properly dealing with the details in motion model and model updater is also the key to good performance. By assembling the basic components properly, we can achieve results comparable with the state of the art without resorting to complicated techniques.  We conclude this paper by highlighting some limitations of our proposed approach as well as some possible ways to address them in our future work.

\section {Related Work}
Significant advances in short-term single-object model-free tracking research have been made over the past few decades.  It is impossible to review them all here due to space limitations.  For a comprehensive survey, readers are referred to~\cite{survey, benchmark2}.

Briefly speaking, there are two major categories of trackers: generative trackers and discriminative trackers.  Generative trackers typically assume a generative process of the appearance of the target and search for the most similar candidate in the video. Some representative methods are (robust) PCA~\cite{ivt, lsst}, sparse coding~\cite{l1t}, and dictionary learning~\cite{onndl}.  On the other hand, discriminative trackers take a different approach.  They usually train a classifier to separate the target from the background.  Thanks to advances made by machine learning researchers, many sophisticated techniques have been applied to visual tracking, including boosting~\cite{oab, semiB}, multiple-instance learning~\cite{mil}, structured output SVM~\cite{struck}, Gaussian process regression~\cite{gpr}, and deep learning~\cite{dlt, sodlt, cnnt}.  Recent benchmarking studies show that the top-performing trackers are usually discriminative trackers~\cite{dsst, kcf} or hybrid ones~\cite{scm} mainly because purely generative trackers cannot handle complicated background well, making it easy to drift away from the target.

As for tracker evaluation, we have witnessed an exploding trend in building datasets and the corresponding benchmarks for visual tracking.  A milestone is the recent contribution made by a benchmark~\cite{benchmark} which consists of 50 videos with full annotations.  The authors also proposed a novel performance metric which uses the area under curve (AUC) of the overlap rate curve or the central pixel distance curve for evaluation.  Recently this benchmark has been extended to an even larger one~\cite{benchmark2}.  Another representative work is the Visual Object Tracking (VOT) challenge~\cite{vot14} which has been held annually since 2013.  The key difference with the benchmark above lies in the evaluation metric.  To characterize better the properties of short-term tracking, evaluation is based on two independent metrics: accuracy and robustness.  While accuracy is measured in terms of the overlap rate between the prediction and ground truth when the tracker does not drift away, robustness is measured according to the frequency of tracking failure which happens when the overlap rate is zero.  Whenever such failure occurs, the tracker is reset to the correct bounding box to continue tracking. Readers are referred to~\cite{metric} for more details.
Other benchmark datasets include the Princeton tracking benchmark~\cite{rgbd}, NUS-PRO~\cite{nuspro} and ALOV++~\cite{survey}.  We tabulate them in Table~\ref{tbl:dataset} for easy comparison.

Another related work is~\cite{bias}.  For fair evaluation of the trackers, the authors first collected evaluation results from the published papers and then removed the results of the proposed method in each paper to reduce subjective bias, because the authors tend to select videos or tune parameters specifically to demonstrate the advantages of the proposed tracker. On the other hand, the authors are usually fair to the other trackers compared . They then used several rank aggregation methods to rank the trackers.  The results are basically consistent with those run directly on the benchmark.

\rowcolors{2}{white}{gray!25}
\begin{table}[htb]
	\begin{center}
	\begin{tabular}{lccc}
		\hline	
		Dataset				&Year		& \#Videos	\\
		\hline
		VTB1.0~\cite{benchmark} 	&2013 	& 50		\\
		PTB~\cite{rgbd}	 		&2013		& 100		\\
		ALOV++ ~\cite{survey}		&2013		& 314		\\
		VOT2014~\cite{vot14} 		&2014		& 25		\\
		VTB2.0~\cite{benchmark2} 	&2015		& 100		\\
		NUS-PRO~\cite{nuspro} 		&2015		& 365		\\
		\hline
	\end{tabular}
	\vspace{2mm}
	\caption{Summary of some visual tracking benchmark datasets.} \label{tbl:dataset}
	\end{center}
\end{table}

\section {Our Proposed Framework}
\begin{figure*}[t]
	\center{
	\includegraphics[width=1.0\linewidth]{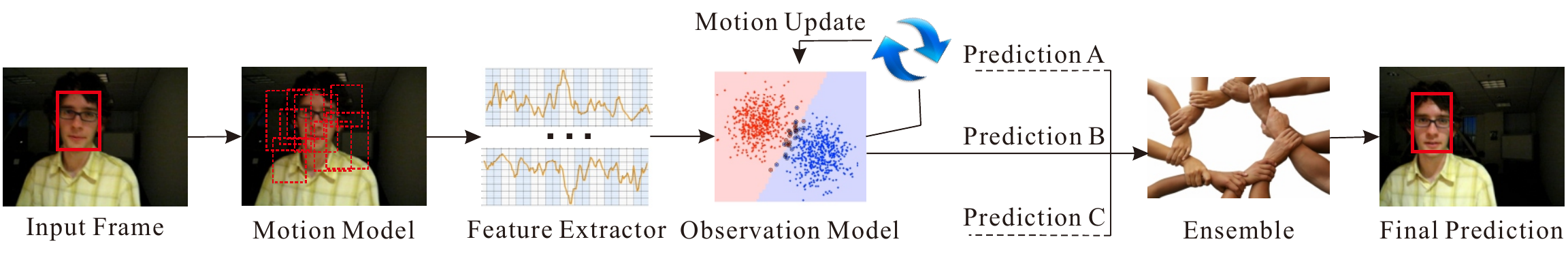}\vspace{-0.2cm}
	\caption{Pipeline of the proposed framework of a visual tracking system.}\label{fig:pipeline}
	}
\end{figure*}
We present our proposed framework in this section.  As mentioned above, we break a tracking system into multiple constituent parts.  Their functions are summarized below:
\begin{enumerate}
\item \textbf{Motion Model:}
Based on the estimation from the previous frame, the motion model generates a set of candidate regions or bounding boxes which may contain the target in the current frame.
\item \textbf{Feature Extractor:}
The feature extractor represents each candidate in the candidate set using some features.
\item \textbf{Observation Model:}
The observation model judges whether a candidate is the target based on the features extracted from the candidate.
\item \textbf{Model Updater:}
The model updater controls the strategy and frequency of updating the observation model.  It has to strike a balance between model adaptation and drift.
\item \textbf{Ensemble Post-processor:}
When a tracking system consists of multiple trackers, the ensemble post-processor takes the outputs of the constituent trackers and uses the ensemble learning approach to combine them into the final result.
\end{enumerate}

A tracking system usually works by initializing the observation model with the given bounding box of the target in the first frame.  In each of the following frames, the motion model first generates candidate regions or proposals for testing based on the estimation from the previous frame.  The candidate regions or proposals are fed into the observation model to compute their probability of being the target.  The one with the highest probability is then selected as the estimation result of the current frame.  Based on the output of the observation model, the model updater decides whether the observation model needs any update and, if needed, the update frequency.  Finally, if there are multiple trackers, the bounding boxes returned by the trackers will be combined by the ensemble post-processor to obtain a more accurate estimate.  This pipeline is illustrated in Fig.~\ref{fig:pipeline}.

\section{Validation Setup}
In this section, we will first introduce our experimental settings which include the dataset and the evaluation metric.  A basic model will then be used as the starting point for illustration.  We plan to make our full implementation publicly available, if the paper is accepted, to facilitate conducting controlled experiments.

\subsection{Settings}
Due to space limitations, we cannot provide in the paper the detailed parameter settings for each component.  Instead, we leave them to the supplemental material.  We determine the parameters of each component using five videos outside the benchmark and then fix the parameters afterwards throughout the evaluation unless specified otherwise.  For this paper, we use the most common dataset, VTB1.0~\cite{benchmark}, as our benchmark.  However, the evaluation approach demonstrated in this paper can be readily applied to other benchmarks as well.

Following the convention of~\cite{benchmark}, we use two metrics for evaluation.  The first one is the AUC of the overlap rate curve.  In each frame, the performance of a tracker can be measured by the overlap rate between the ground-truth and predicted bounding boxes, where the overlap rate is defined as the area of intersection of the two bounding boxes over the area of their union.  With a given threshold for the overlap rate, we can calculate the success rate of the tracker over all the video frames.  By varying the threshold from 0 gradually to 1, it will yield a curve which varies from it maximum successful rate to success rate 0 accordingly.  A larger AUC of this curve indicates a higher accuracy of the tracker.  The second metric is the precision at threshold 20 for the central pixel error curve.  The curve is generated in a way similar to that for the overlap rate.  The central pixel error is defined as the distance between the centers of the two bounding boxes in pixels.
This metric is useful for the cases that the scale of the object changes but the tracker does not support scale variation, since using only the scale of the first frame will definitely give a low overlap rate which will make the results indistinguishable.


\subsection{Basic Model}
\begin{figure}[htb]
	\centering
	\includegraphics[width=0.49\linewidth]{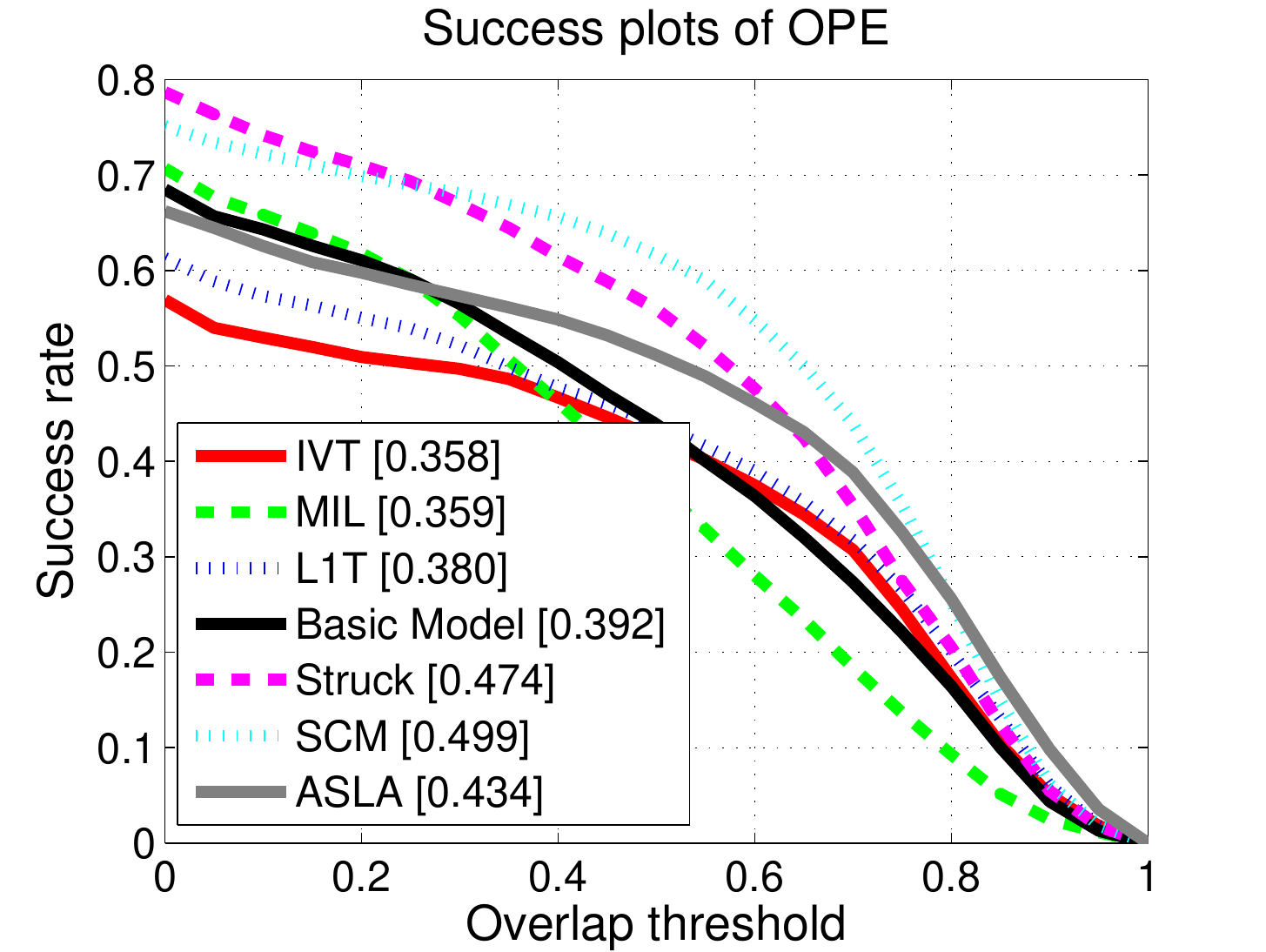}
	\includegraphics[width=0.49\linewidth]{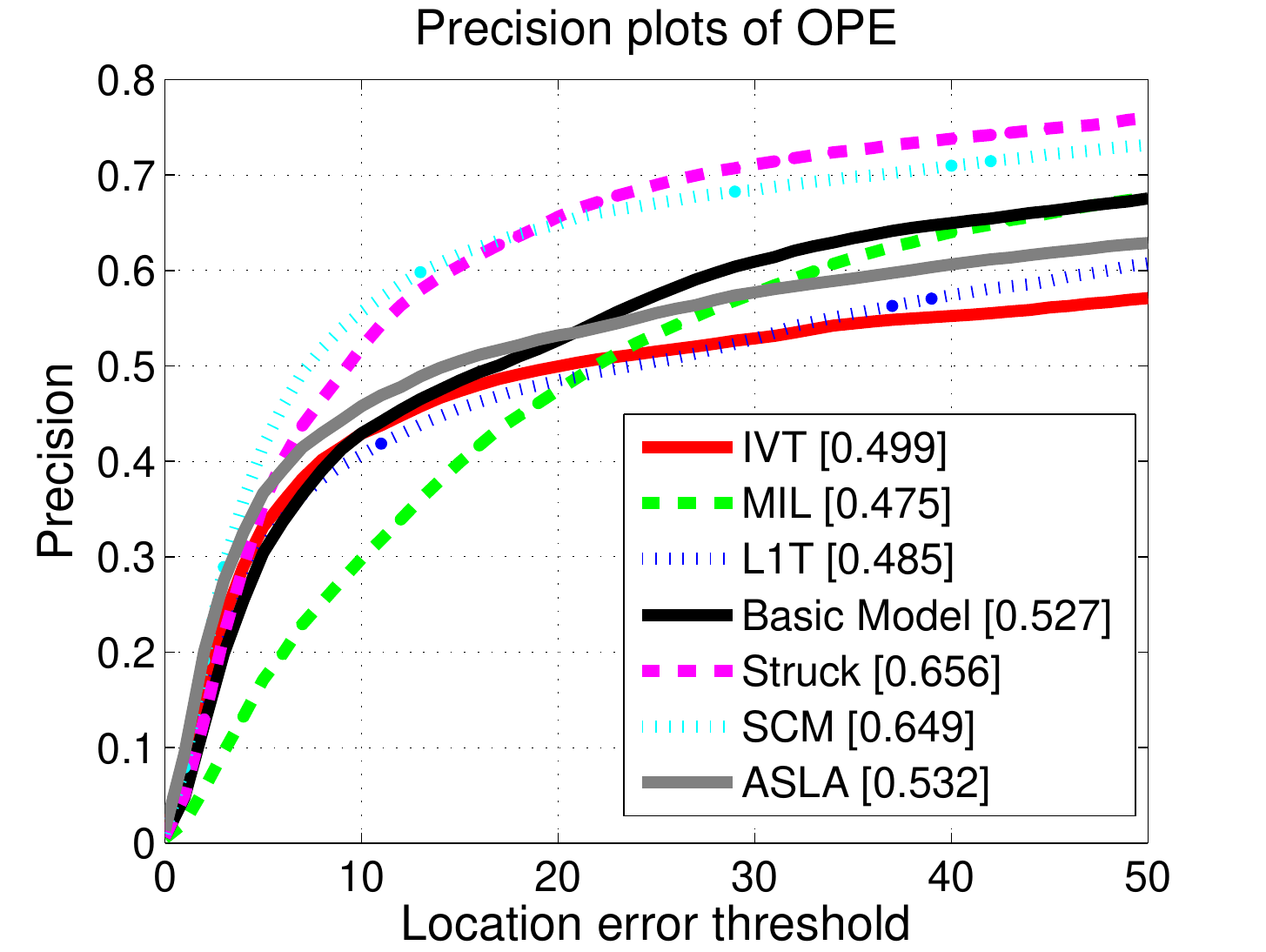}
	\caption{\label{fig:overall_OPE}One Pass Evaluation (OPE) plots on VTB1.0~\cite{benchmark}. The performance  score for each tracker is shown in the legend. For the success plots of  overlapping rate, the score is the AUC value. While for precision plots of central pixel error, the score is the precision at threshold 20.  }
	\label{fig:basicModel}
\end{figure}

We need a basic model to start our analysis.  As a starting point, we use a very simple one which adopts the particle filter framework as the motion model, raw pixels of grayscale images as features, and logistic regression as the observation model.  For the model updater, we use a simple rule that if the highest score among the candidates tested is below a threshold, the model will be updated.
Moreover, we only consider a single tracker in this basic model and hence no ensemble post-processor will be used.  Details of all these components will be provided in the next section.  For illustration, we show in Fig.~\ref{fig:basicModel} the performance of this basic model along with some popular trackers.  We can see that even this very simple model can obtain moderate results when compared to some competitive methods in~\cite{benchmark}.

\section{Validation and Analysis}
We now conduct an ablative analysis to see how each component of a tracker affects its final tracking performance.  We present our analysis of different components in the order of their importance and necessity.

\subsection{Feature Extractor}
The feature extractor converts the raw image data into some (usually) more informative representation.  Five feature representations are commonly used for object detection and tracking:
\begin{enumerate}
	\item \textbf{Raw Grayscale:} It simply resizes the image into a fixed size, converts it to grayscale, and then uses the pixel values as features.
	\item \textbf{Raw Color:} It is the same as raw grayscale features except that the image is represented in the CIE Lab color space instead of grayscale.
	\item \textbf{Haar-like Features:} We consider the simplest form, rectangular Haar-like features, which was first introduced in 2001~\cite{haar}.
	\item \textbf{HOG:} It is a good shape detector widely used for object detection. It was first proposed in 2005~\cite{hog}.
	\item \textbf{HOG + Raw Color:} This feature representation simply concatenates the HOG and raw color features.
\end{enumerate}

\begin{figure}[htb]
	\centering
	\includegraphics[width=0.49\linewidth]{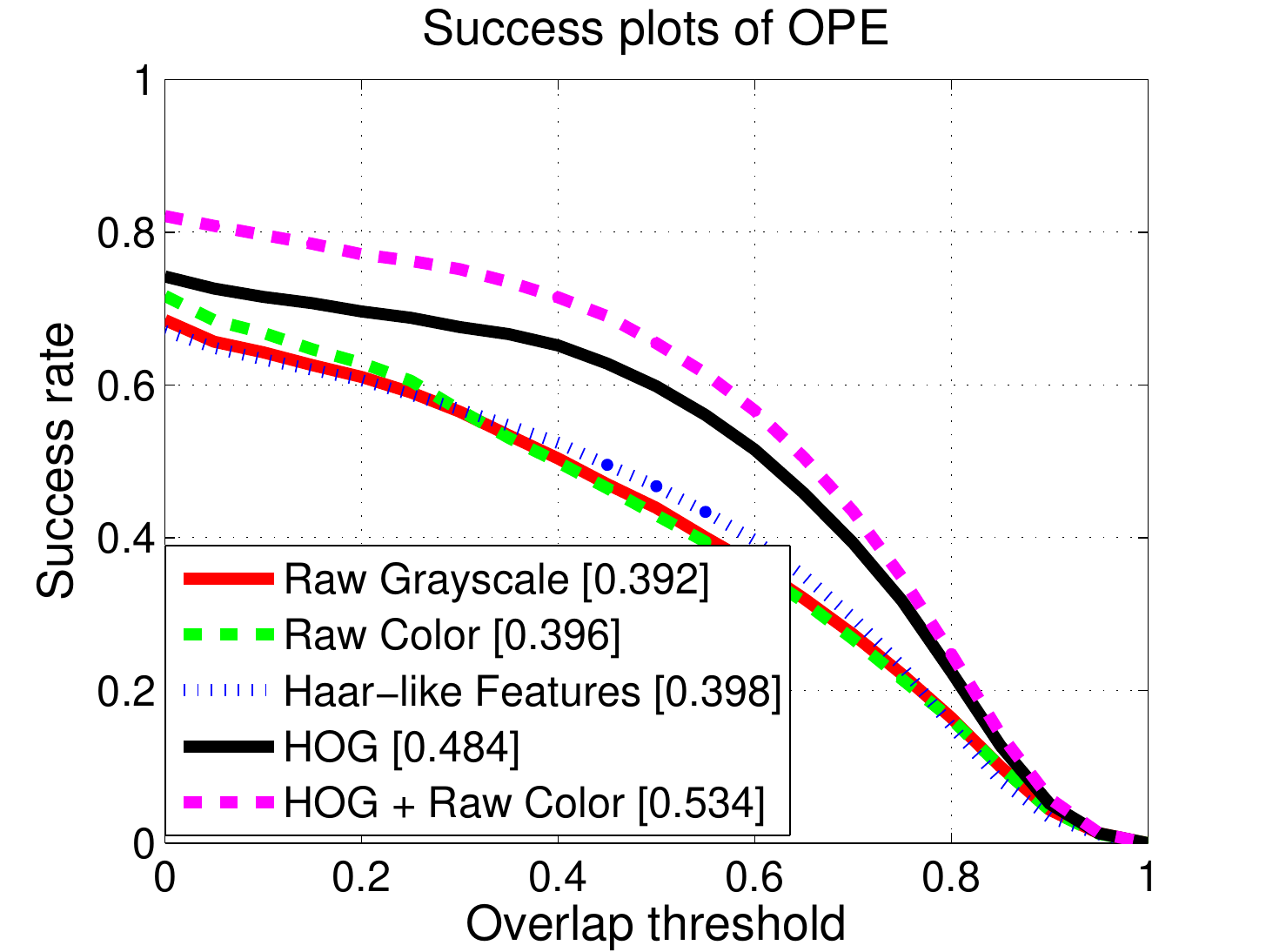}
	\includegraphics[width=0.49\linewidth]{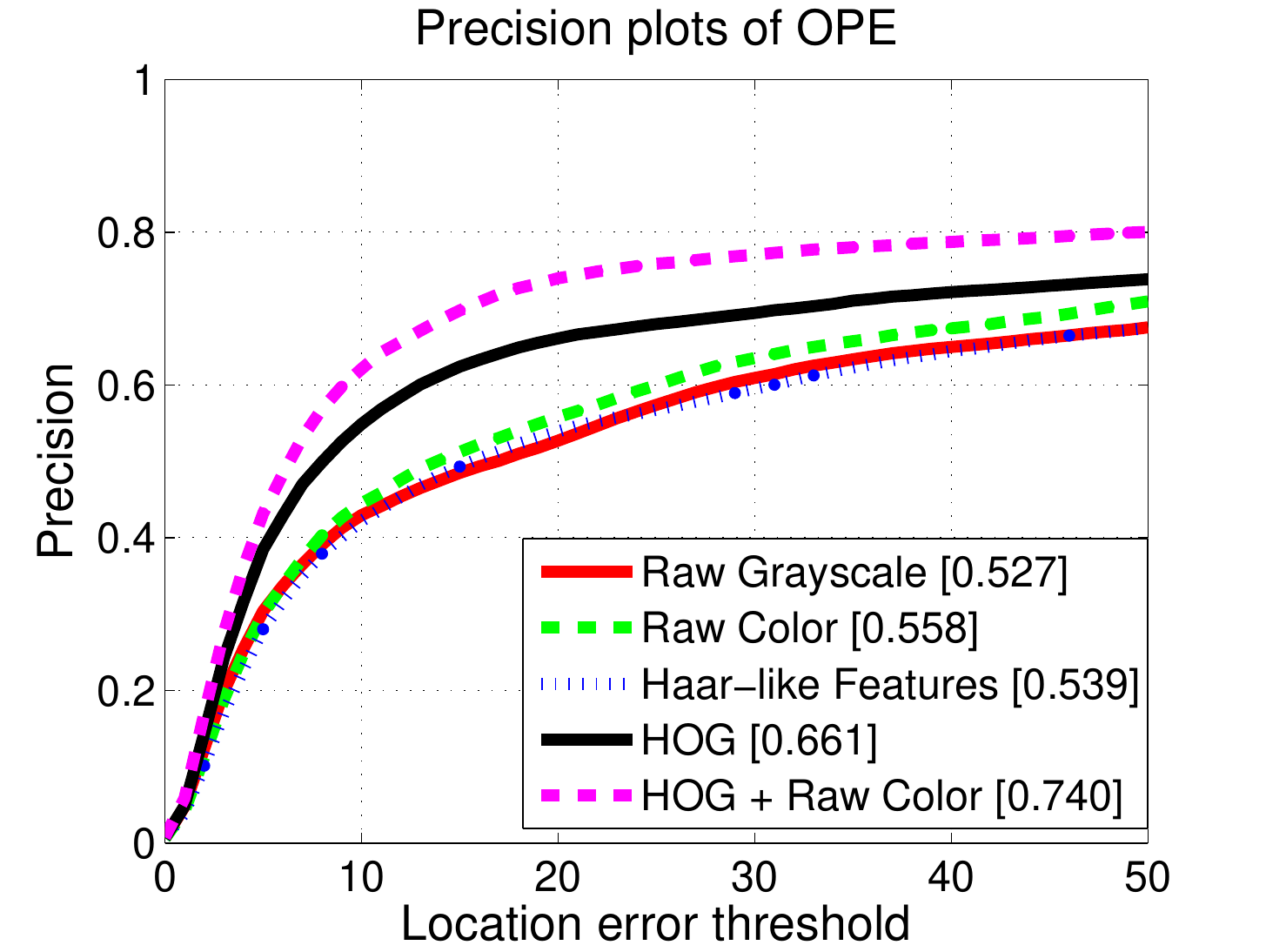}
	\caption{\label{fig:overall_OPE} Results of different feature representations.}
	\label{fig:feature}
\end{figure}

We compare the performance of these feature representations in Fig.~\ref{fig:feature}.
Note that the performance gaps between features can be quite large.  For example, the best scheme (HOG + raw color) outperforms the basic model (raw grayscale) by more than $20\%$.  In fact, the best result is even beyond the best performance reported in~\cite{benchmark}.  Although there exist even more powerful features such as those extracted by the convolutional neural network (CNN) and they indeed can yield state-of-the-art performance~\cite{sodlt, cnnt}, na\"{i}ve application of this approach will incur high computational cost which is highly undesirable for tracking applications.  For efficiency consideration, some special designs as in~\cite{sodlt} are needed.  Another interesting direction is to exploit the color information.  Some recent methods~\cite{adaptiveColor, defenseColor} demonstrated notable performance with carefully designed color features.  Not only are these features lightweight, but they are also suitable for deformable objects.  We believe that finding good features for object tracking is still a research direction that is worth pursuing.

\paragraph{Our Findings:} \emph{The feature extractor is the most important component of a tracker. Using proper features can dramatically improve the tracking performance. Developing a good and effective feature representation for tracking is still an open problem.}

\subsection{Observation Model}
\begin{figure}[htb]
	\centering
	\includegraphics[width=0.48\linewidth]{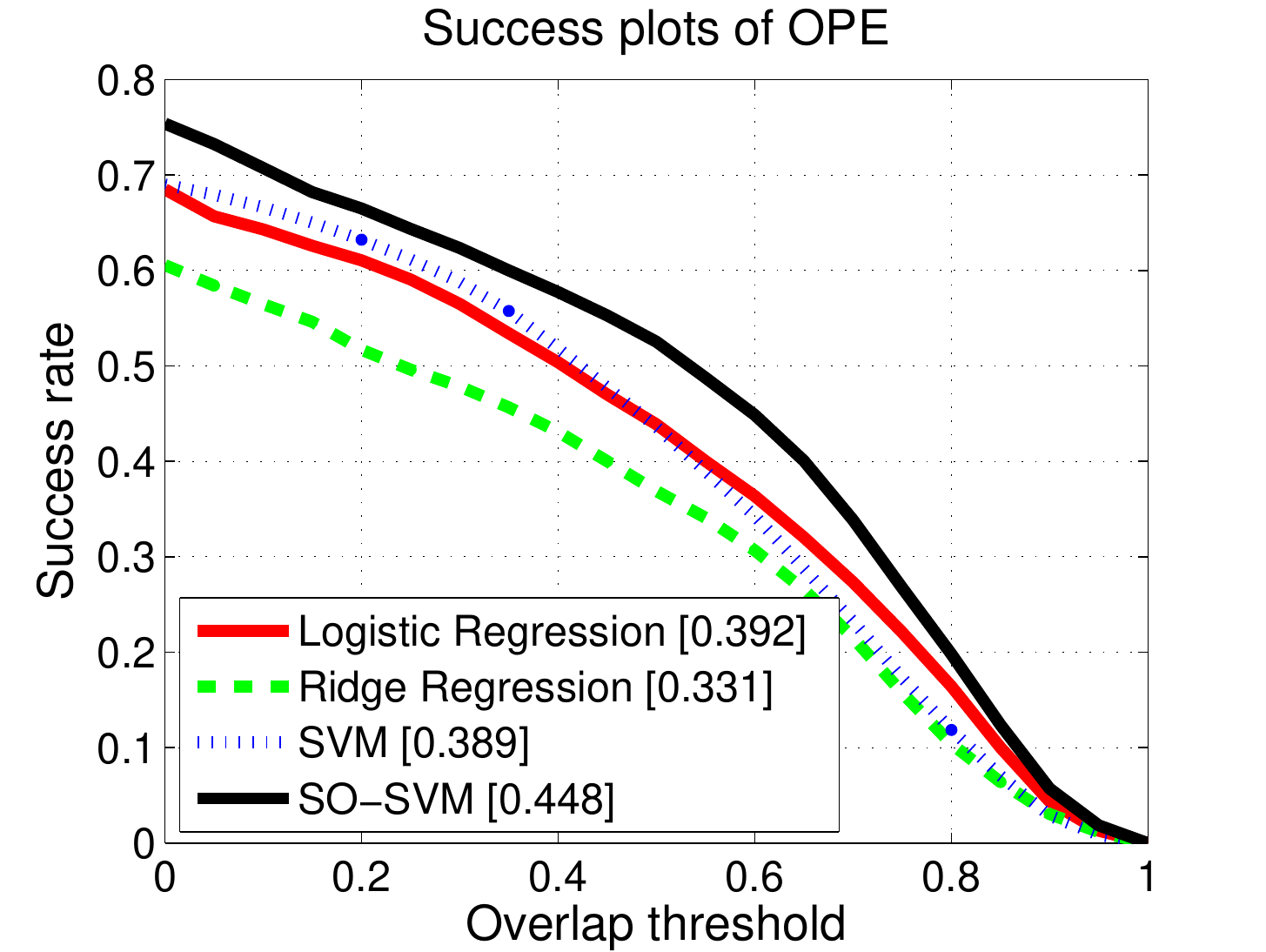}
	\includegraphics[width=0.48\linewidth]{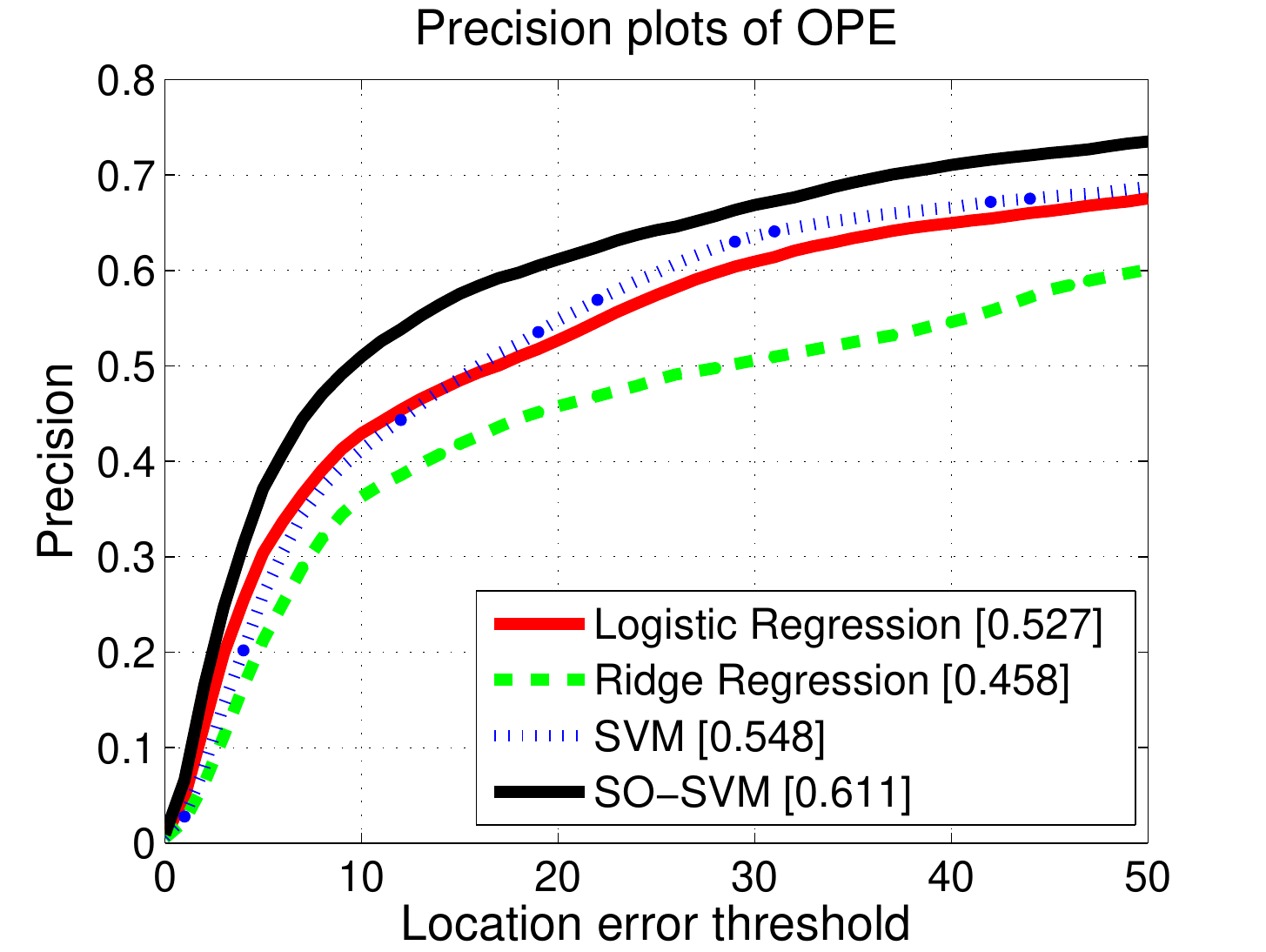}
	\caption{ Results of different observation models with weak features.}
	\label{fig:obs_gray}
\end{figure}

\begin{figure}[htb]
	\centering
	\includegraphics[width=0.48\linewidth]{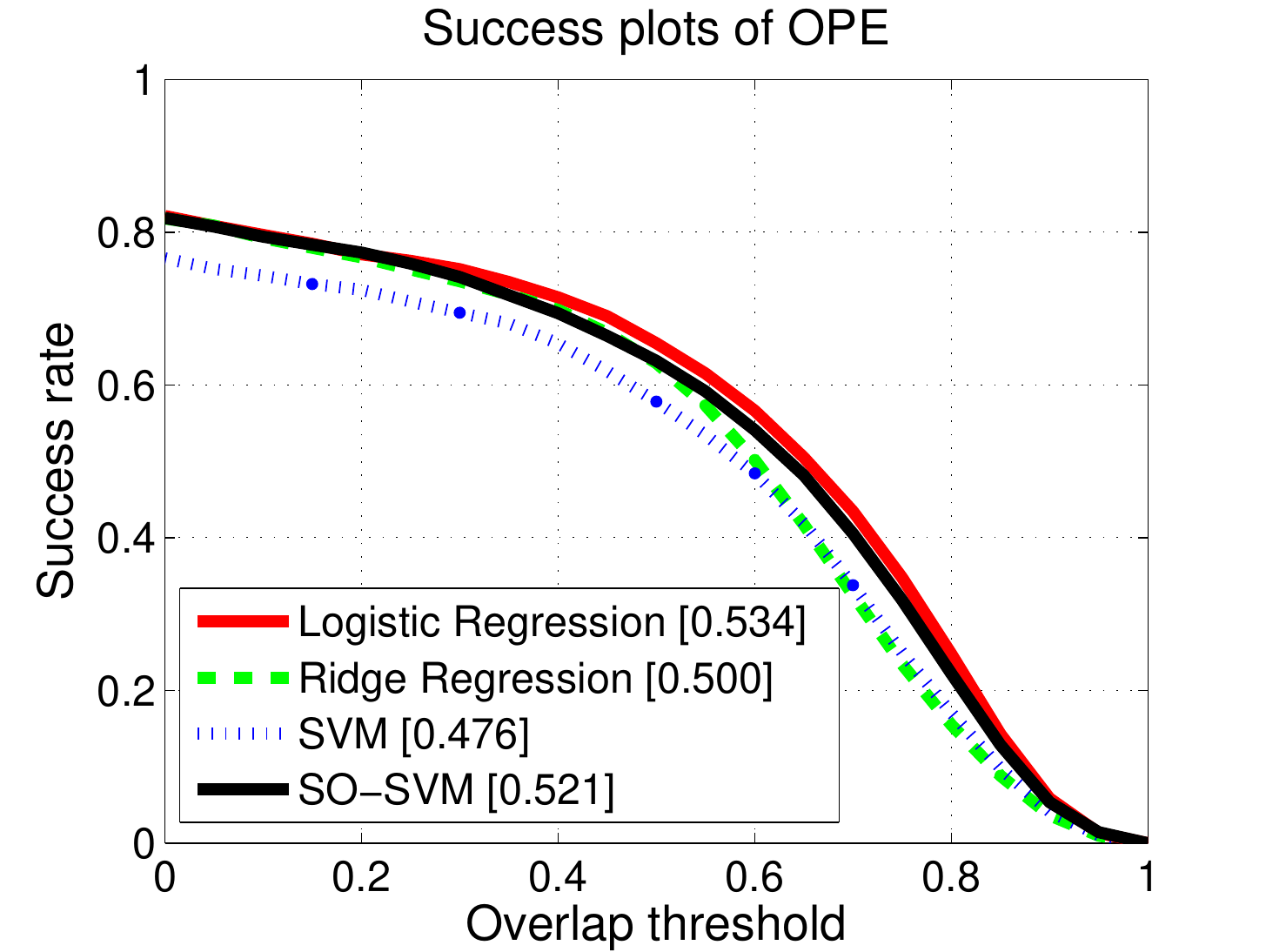}
	\includegraphics[width=0.48\linewidth]{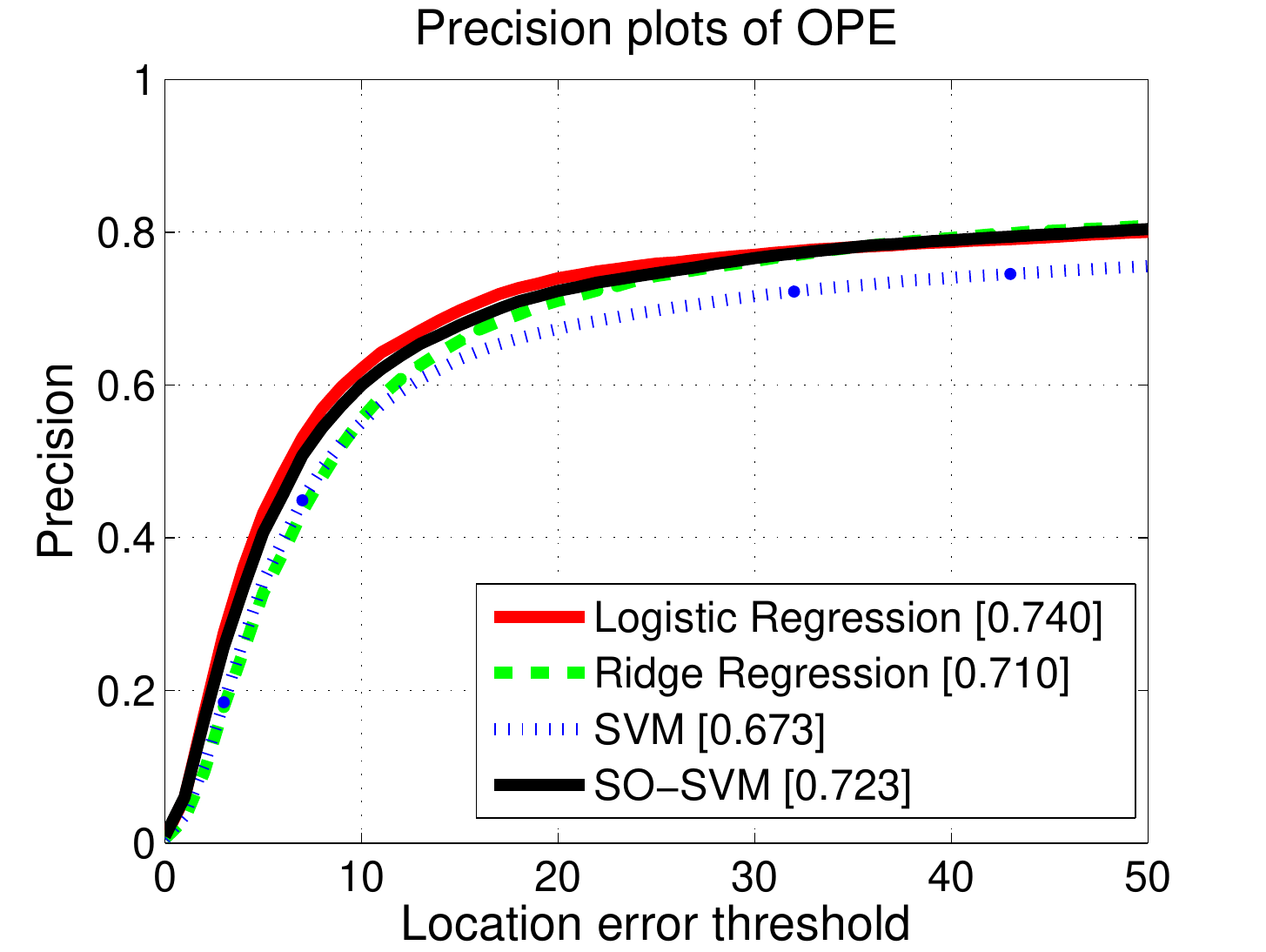}
	\caption{ Results of different motion models with strong features.}
	\label{fig:obs_hograw}
\end{figure}

The observation model returns the confidence of a given candidate being the target, so it is usually believed to be the key component of a tracker.  Since the top-performing trackers in recent benchmarking studies are exclusively discriminative trackers, we do not include generative observation models in our analysis.  We consider the following observation models:
\begin{enumerate}
	\item \textbf{Logistic Regression:} Logistic regression with $l_2$ regularization is used. Online update is achieved by simply using gradient descent.
	\item \textbf{Ridge Regression:} Least squares regression with $l_2$ regularization is used. The targets for positive examples are set to one while those for negative examples are set to zero.  Online update is achieved by aggregating sufficient statistics, a scheme originated from~\cite{onlineDL} for online dictionary learning.
	\item \textbf{SVM:} Standard SVM with hinge loss and $l_2$ regularization is used. The online update method is from~\cite{onlinesvm}.
	\item \textbf{Structured Output SVM (SO-SVM):} The optimization target of the structured output SVM is the overlap rate instead of the class label. This method is from~\cite{struck}.
\end{enumerate}
We test these four classifiers using two feature representations, a weak one (raw grayscale) and a strong one (HOG + raw color).  The results are shown in Fig.~\ref{fig:obs_gray} and Fig.~\ref{fig:obs_hograw}, respectively.

When weak features are used, a powerful classifier such as SO-SVM can indeed improve the performance of the basic model by about $10\%$.  However, when strong features are used, surprisingly the results are reversed.  Logistic regression becomes the best-performing observation model.  Similar observation was also reported in~\cite{kcf}: when raw pixels are used as features, a kernelized classifier beats a simple linear one by a large margin; however, when HOG features are used, the performance gap reduces to almost zero.  We believe that our finding is by no means just coincidence.

\paragraph{Our Findings:} \emph{Different observation models indeed affect the performance when the features are weak. However, the performance gaps diminish when the features are strong enough. Consequently, satisfactory results can be obtained even using simple classifiers from textbooks.}

\subsection{Motion Model}
\begin{figure}[htb]
	\centering
	\includegraphics[width=0.48\linewidth]{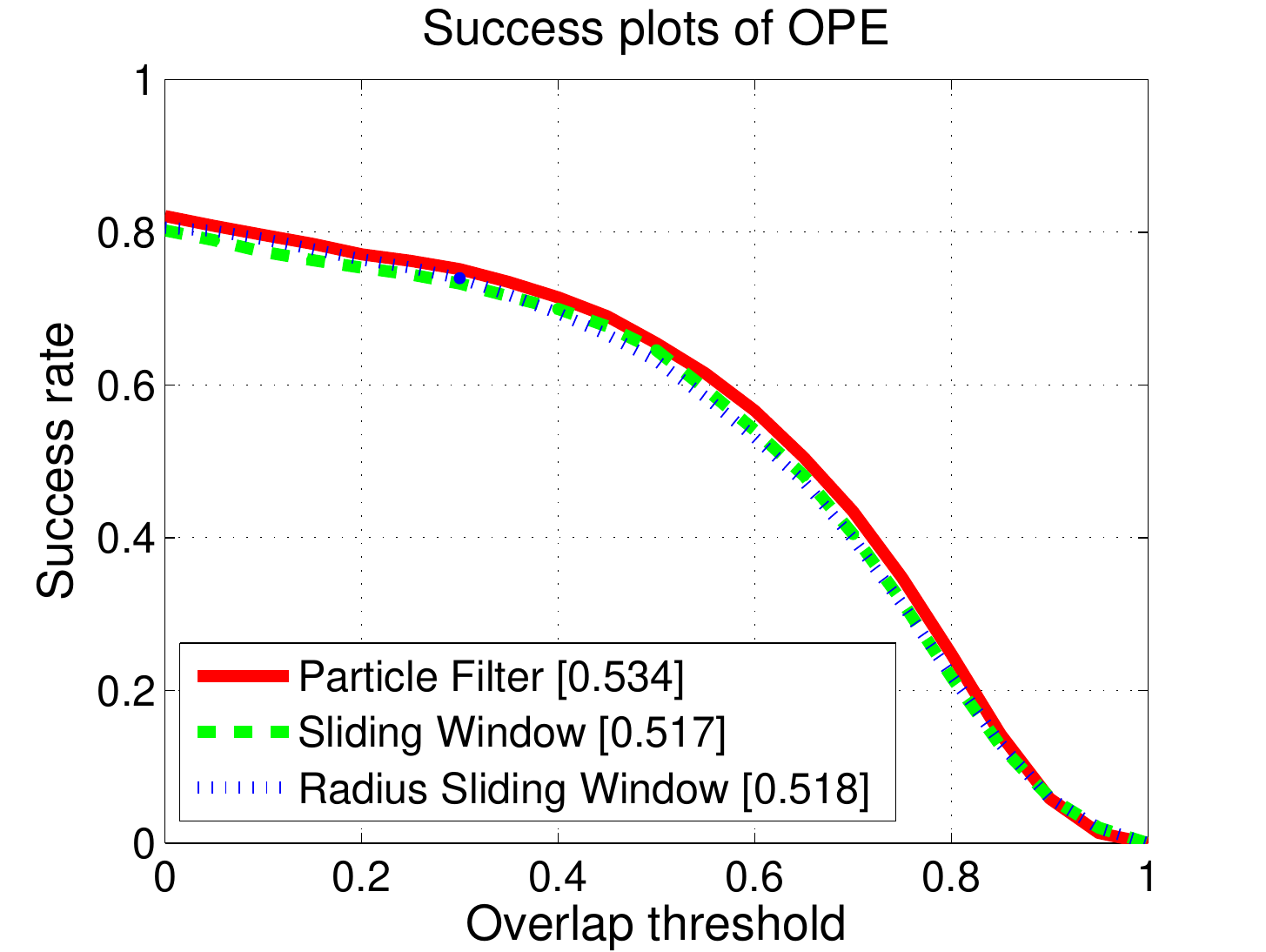}
	\includegraphics[width=0.48\linewidth]{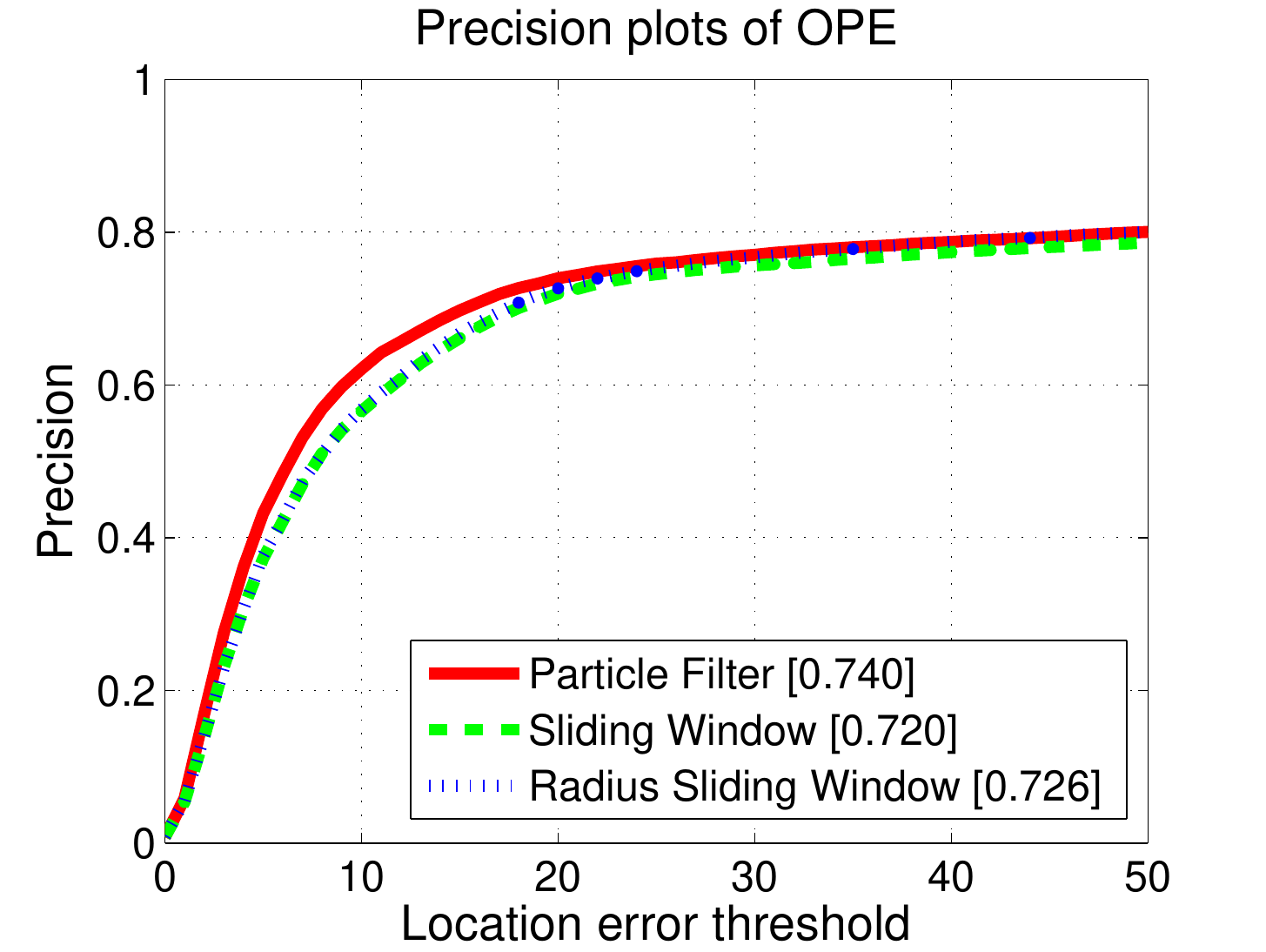}
	\caption{Results of different motion models.}\vspace{-0.1cm}
	\label{fig:motion_overall}
\end{figure}

In each frame, based on the estimation from the previous frame, the motion model generates a set of candidates for the target.  We consider three commonly used motion models:
\begin{enumerate}
	\item \textbf{Particle Filter:} Particle filter is a sequential Bayesian estimation approach which recursively infers the hidden state of the target. For a complete tutorial, we refer the readers to~\cite{tutorial} for details.
	\item \textbf{Sliding Window:} The sliding window approach is an exhaustive search scheme which simply considers all possible candidates within a square neighborhood.
	\item \textbf{Radius Sliding Window:} It is a simple modification of the previous approach which considers a circular region instead. It was first considered in~\cite{struck}.
\end{enumerate}
The key differences between the particle filter and sliding window approaches lie in the following two aspects.  First, the particle filter approach can maintain a probabilistic estimation for each frame.  Thus when several candidates have high probability of being the target, they will all be kept for the next frames.  As a result, it can help to recover from tracker failure.  In contrast, the sliding window approach only chooses the candidate with the highest probability and prune all others.  Second, the particle filter framework can easily incorporate changes in scale, aspect ratio, and even rotation and skewness.  Due to the high computational cost induced by exhaustive search, however, the sliding window approach can hardly pursue it.  Results of the comparison are shown in Fig.~\ref{fig:motion_overall}.

We note that the three motion models show no significant difference on the benchmark.  Although particle filter has the two advantages mentioned above, they do not translate into performance gain in the evaluation.  Nevertheless, we should note that this observation is valid only when performing object tracking under normal scenarios.  In case there is severe camera shake such as in egocentric videos, more sophisticated motion models specially designed for a purpose are definitely worth trying.

\begin{figure}[htb]
	\centering
	\includegraphics[width=0.495\linewidth]{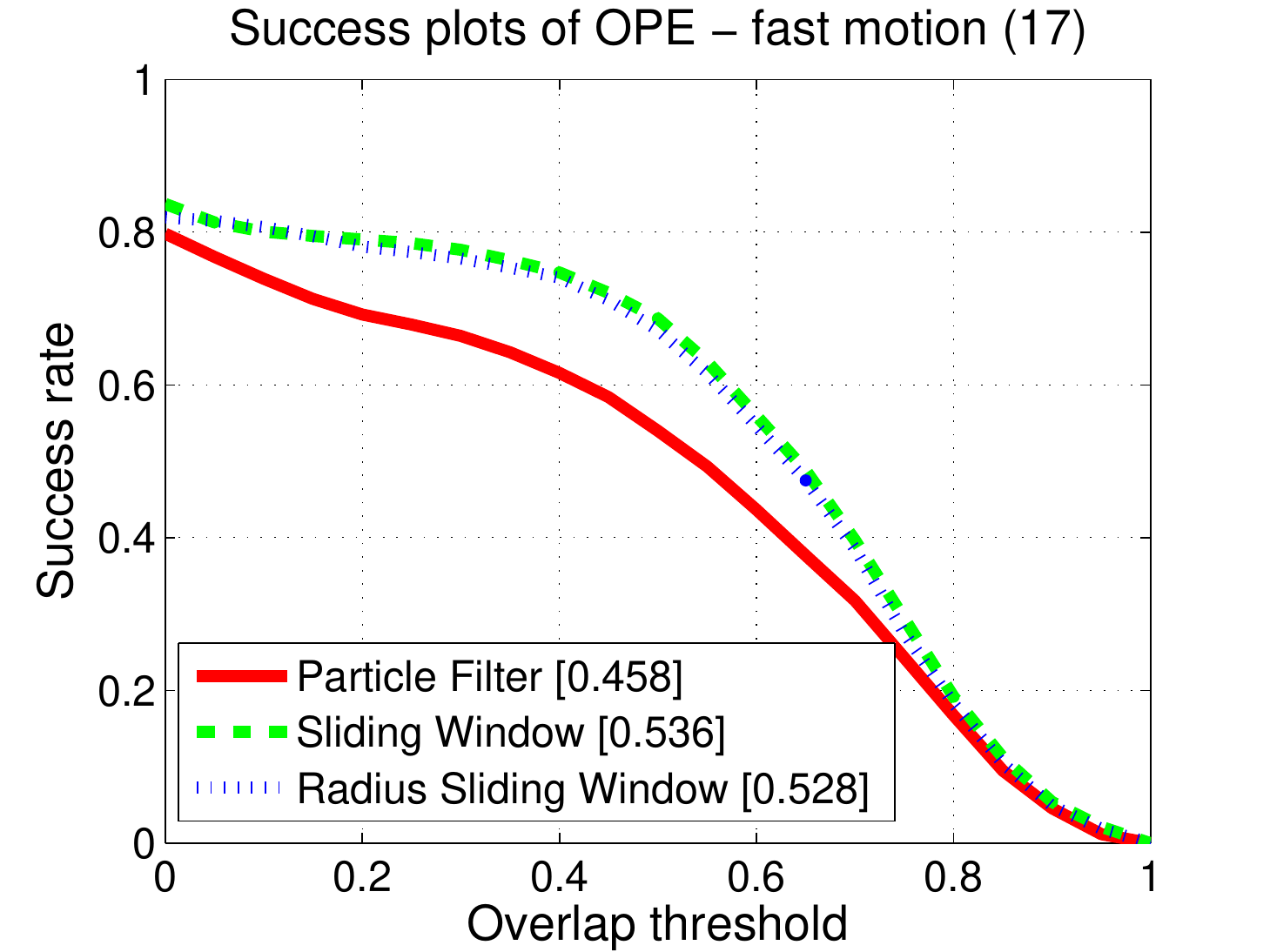}
	\includegraphics[width=0.495\linewidth]{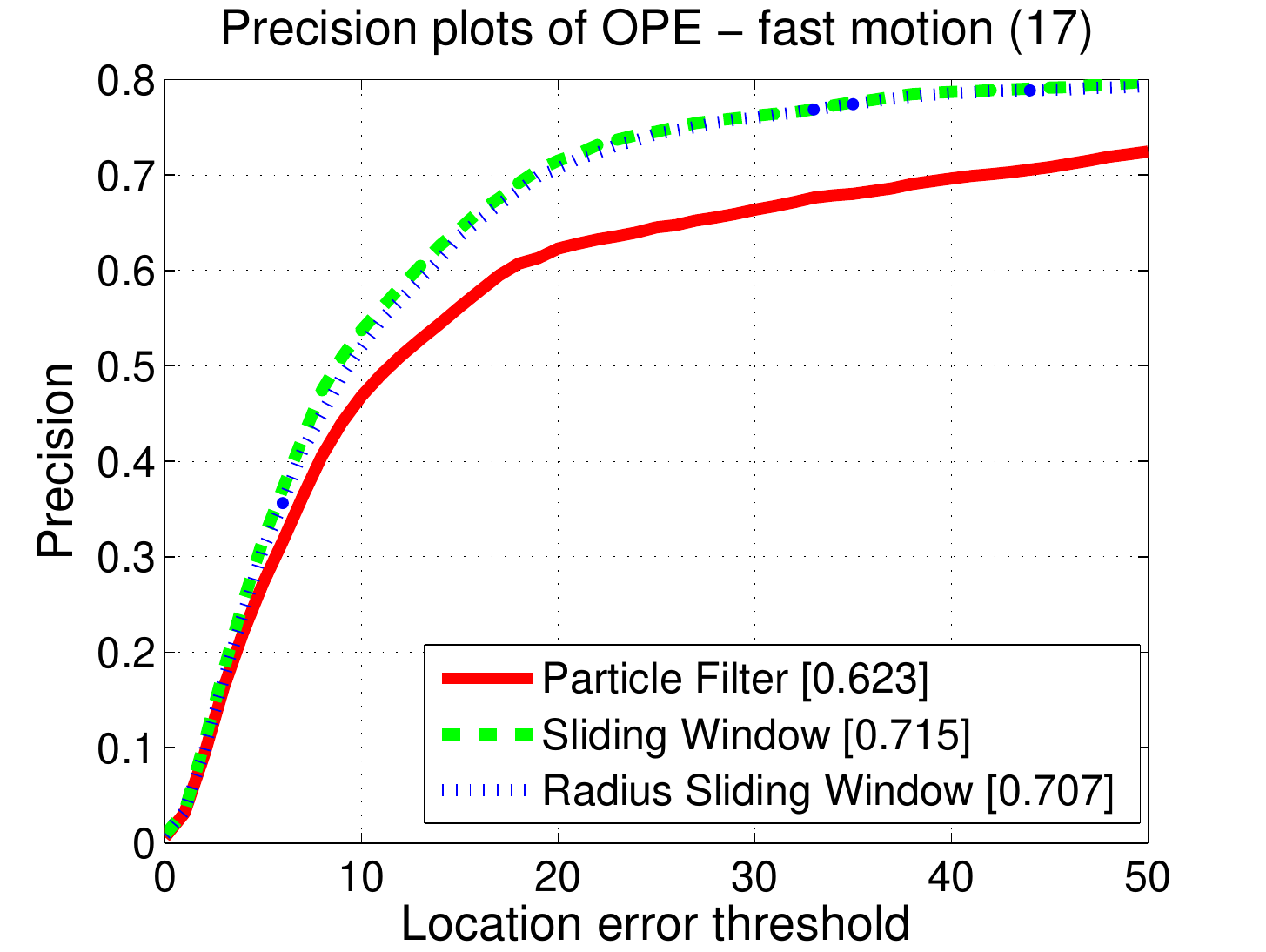}\vspace{0.2cm}\\
	\includegraphics[width=0.495\linewidth]{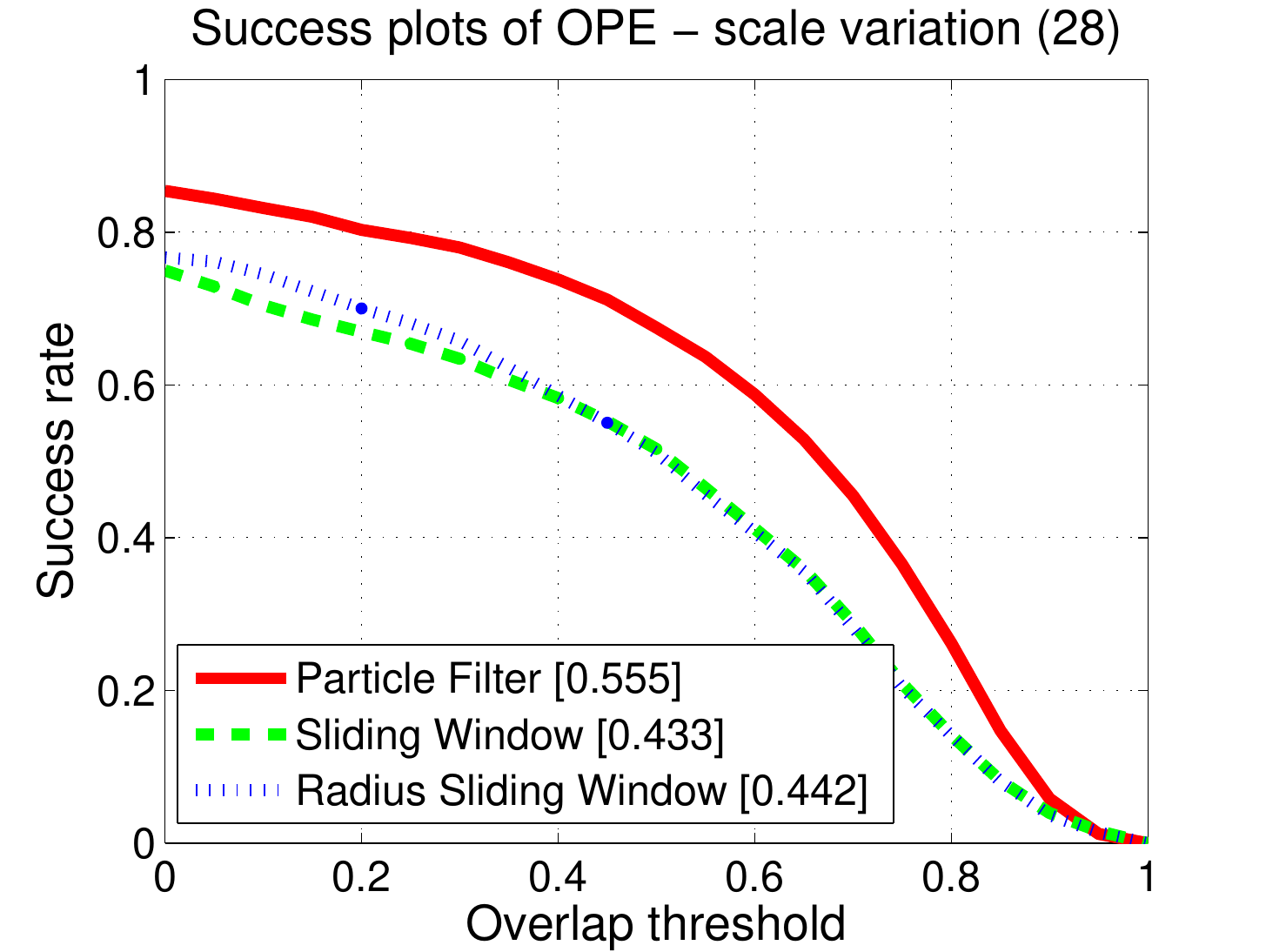}
	\includegraphics[width=0.495\linewidth]{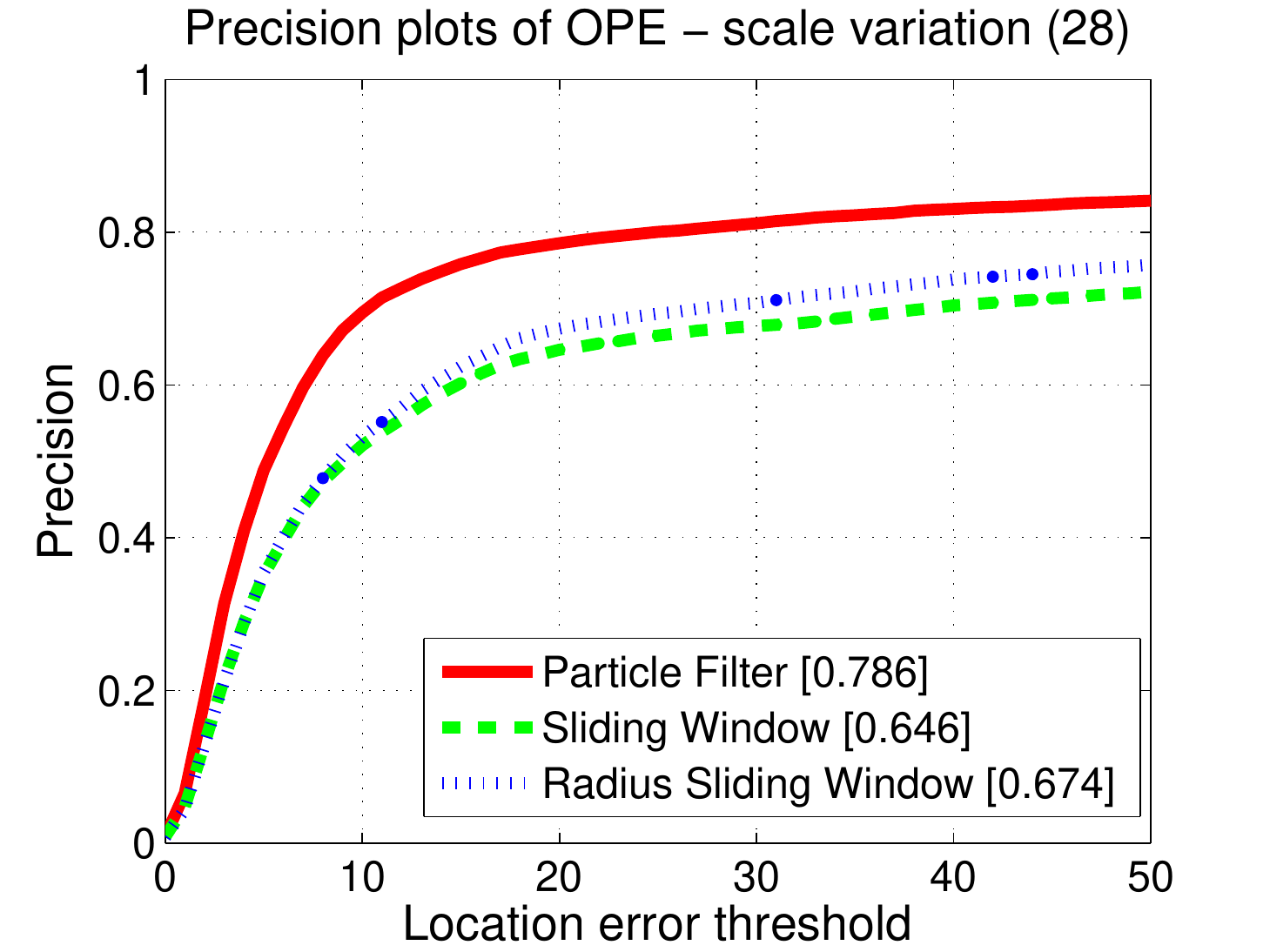}
	\caption{Results of different motion models with fast motion and scale variation.}\vspace{-0.2cm}
	\label{fig:motion_analysis}
\end{figure}

A closer look at the subcategory results of the benchmark in Fig.~\ref{fig:motion_analysis} reveals some interesting observations.
Not surprisingly, particle filter is much better than the sliding window approach when scale variation exists, but it is much worse for the fast motion sub-category.  So, can we perform well in both subcategories simultaneously?

\begin{figure}[htb]
	\centering
	\includegraphics[width=0.495\linewidth]{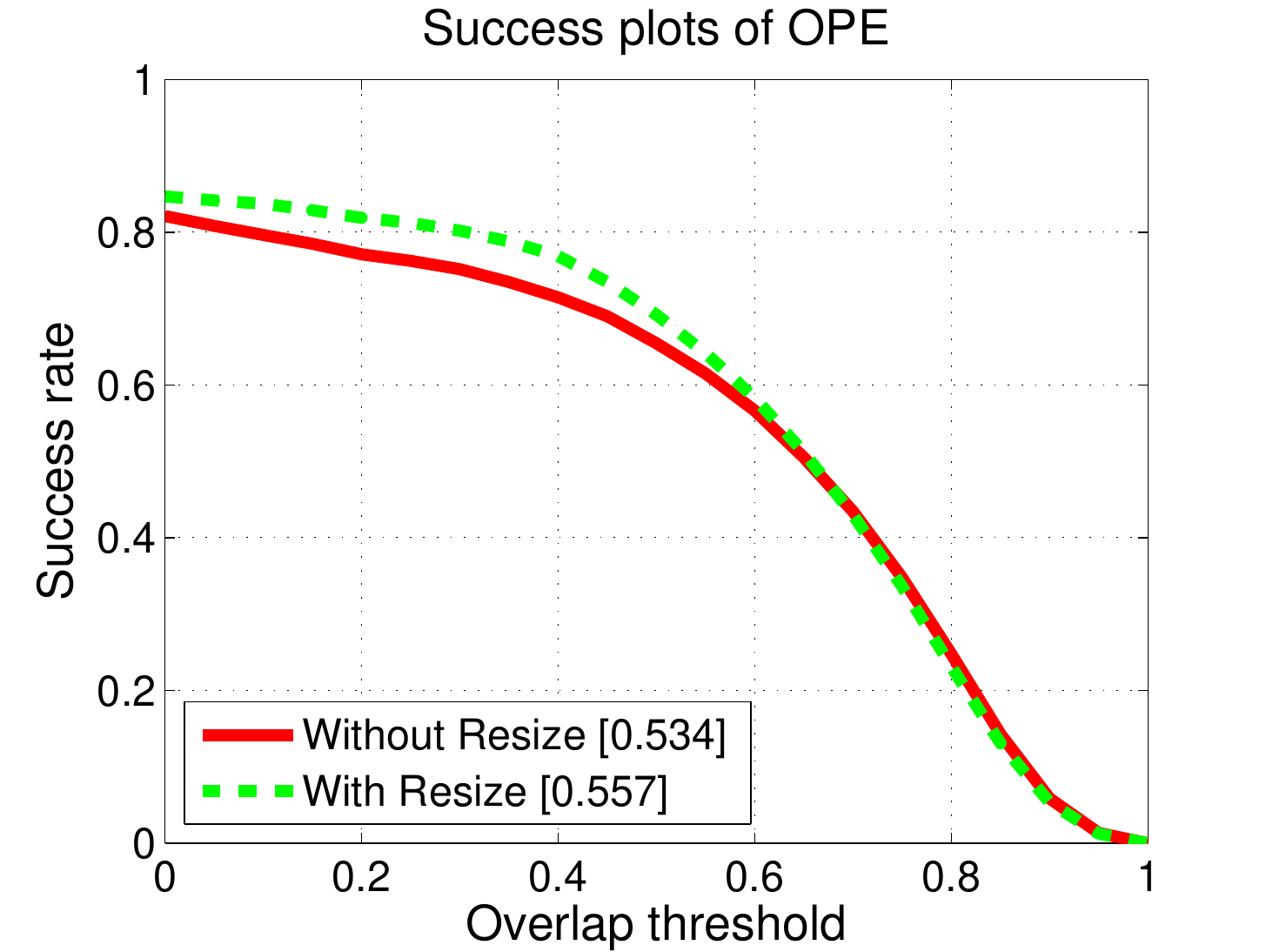}
	\includegraphics[width=0.495\linewidth]{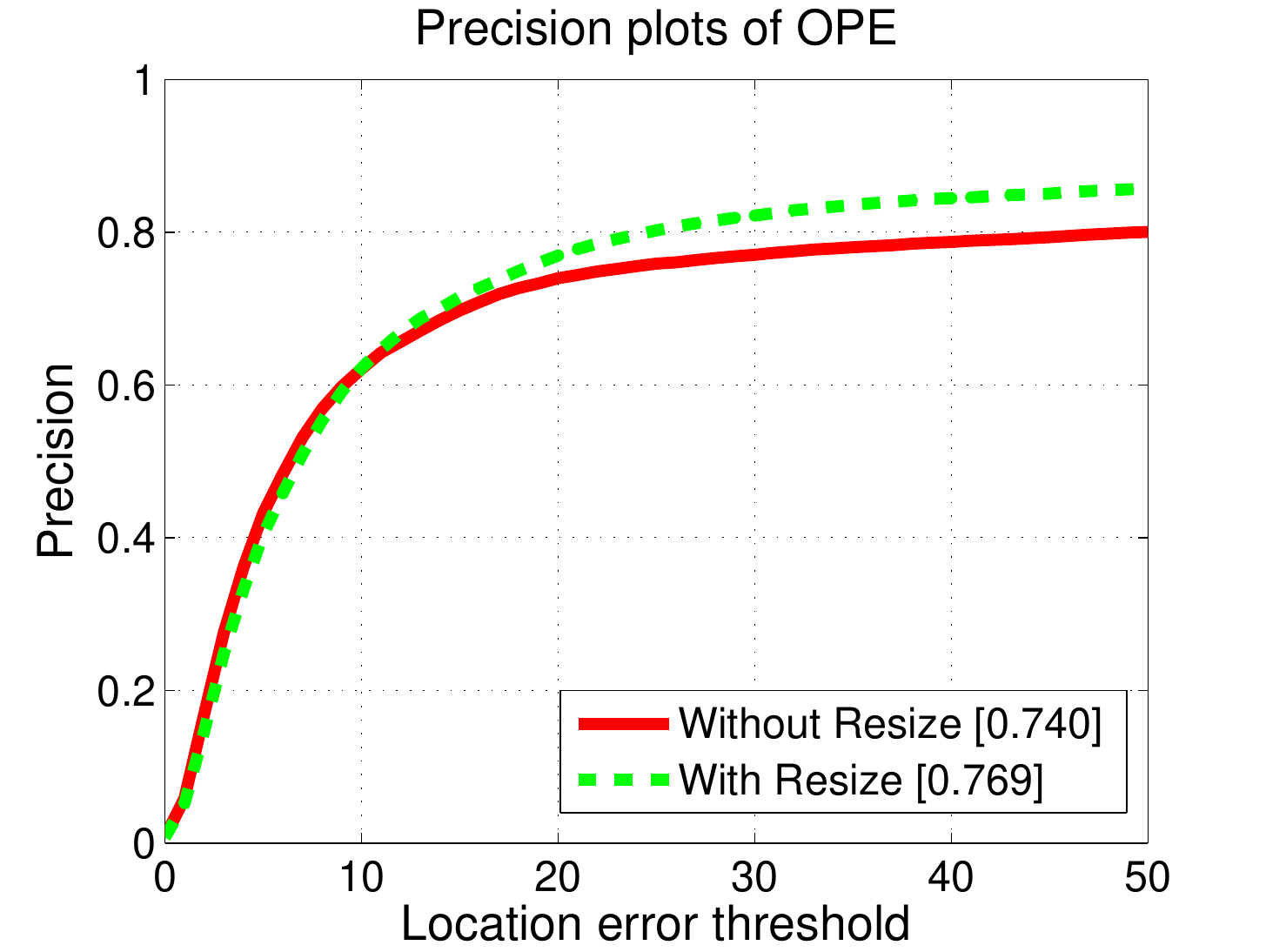}\vspace{0.2cm}\\
	\includegraphics[width=0.495\linewidth]{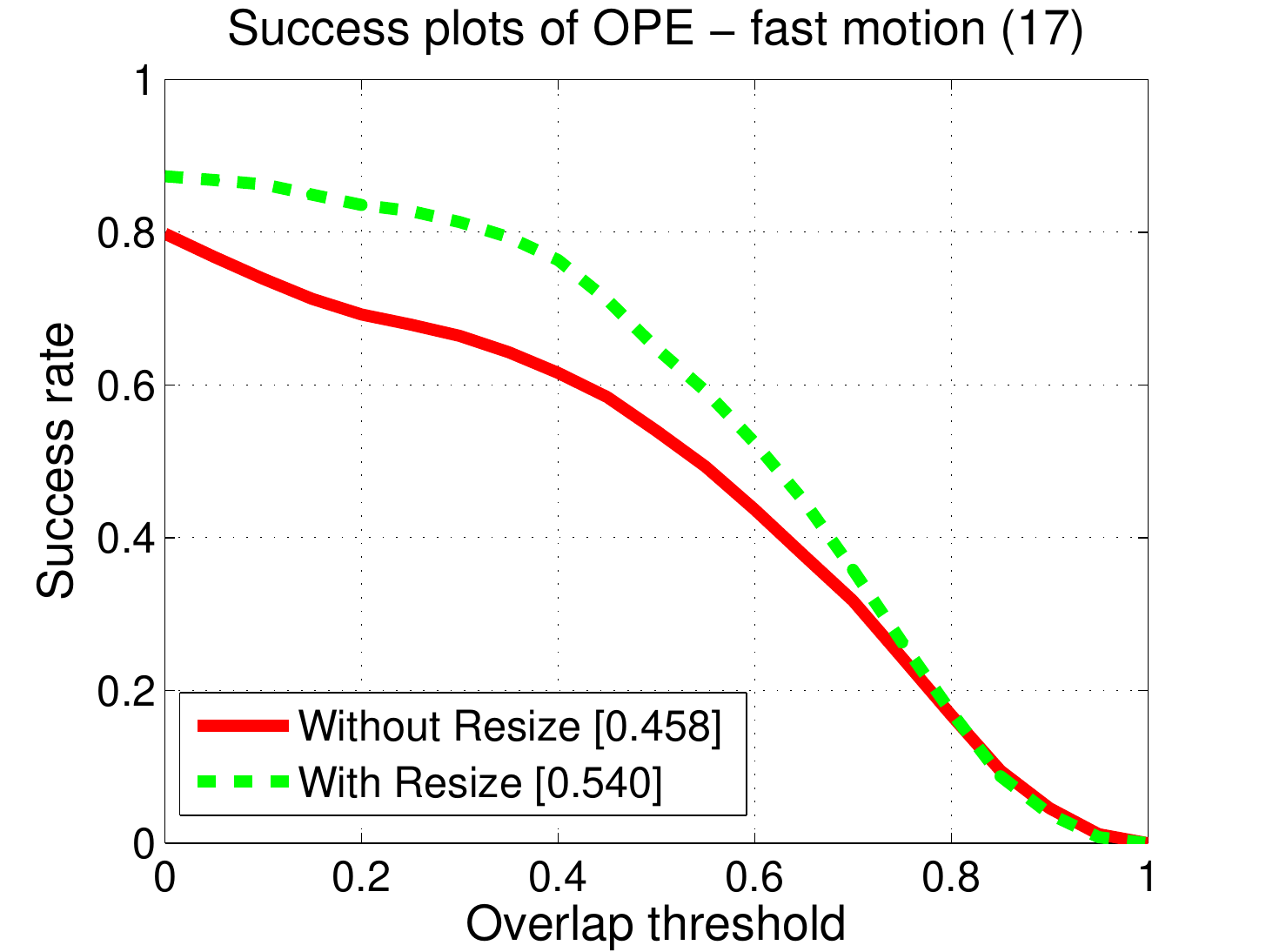}
	\includegraphics[width=0.495\linewidth]{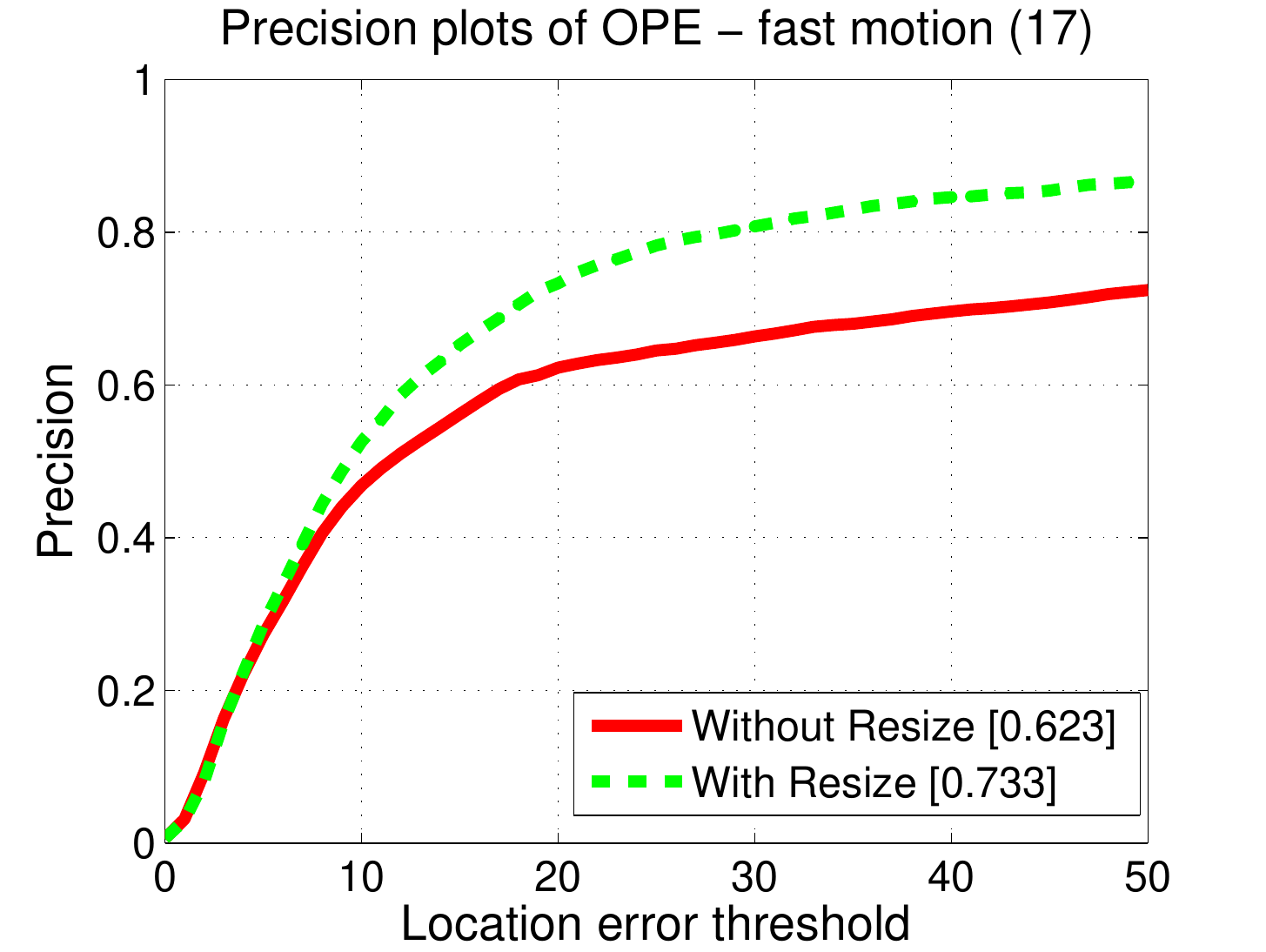}
	\caption{Results comparing the settings with and without resizing the input video to a fixed size.}\vspace{-0.3cm}
	\label{fig:motion_resize}
\end{figure}

To answer this question, we first examine the role of the translation parameters in a particle filter: They control the search region of the tracker. When the search region is too small, the tracker is likely to lose the target when it is in fast motion. On the other hand, having a large search region will make the tracker prone to drift due to distractors in the background.  We have noticed an improper practice in setting the parameters, which is often to use the number of pixels as unit.  However, different videos may have very different resolution.  Using an absolute number of pixels to set the parameters will actually result in different search regions.  A simple solution is to scale the parameters by the video resolution which, equivalently, resizes the video to some fixed scale.  We adopt the latter approach and report the results in Fig.~\ref{fig:motion_resize}.

We find that even such a simple normalization step can improve the performance significantly especially when there exists fast motion. By applying this simple normalization step, particle filter could handle both scale variation and fast motion well. This experiment thus validates our hypothesis that the parameters of the motion model should be adaptive to video resolution.

\vspace{-0.2cm}
\paragraph{Our Findings:} \emph{The motion model only has minor effect on the performance. Nevertheless, setting the parameters properly is still crucial to obtaining good performance.  Due to its ability to adapt to scale changes which are not uncommon in practice, we will still take the particle filter approach with resized input as the default motion model in the sequel.}

\subsection{Model Updater}
The model updater determines both the strategy and frequency of model update.  Since the update of each observation model is different, the model updater often specifies when model update should be done and its frequency.
As under our tracking setting there is only one reliable example, the tracker must maintain a tradeoff between adapting to new but possibly noisy examples collected during tracking and preventing the tracker from drifting to the background.

When the model needs update, we first collect some positive examples whose centers are within 5 pixels from the target and some negative examples within 100 pixels but with overlapping rate less than 0.3.  We consider two model update methods:
\begin{enumerate}
	\item The first method is to update the model whenever the confidence of the target falls below a threshold.  Doing so ensures that the target always has high confidence. This is the default updater used in our basic model.
	\item The second method is to update the model whenever the difference between the confidence of the target and that of the background examples is below a threshold. This strategy simply maintains a sufficiently large margin between the positive and negative examples instead of forcing the target to have high confidence. It is potentially helpful when the target is occluded or disappears. This method was proposed and evaluated in ~\cite{spllt}.	
\end{enumerate}

\begin{figure}[htb]
	\centering\vspace{-0.1cm}
	\subfigure[AUC of overlap rate]{\includegraphics[width=0.48\linewidth]{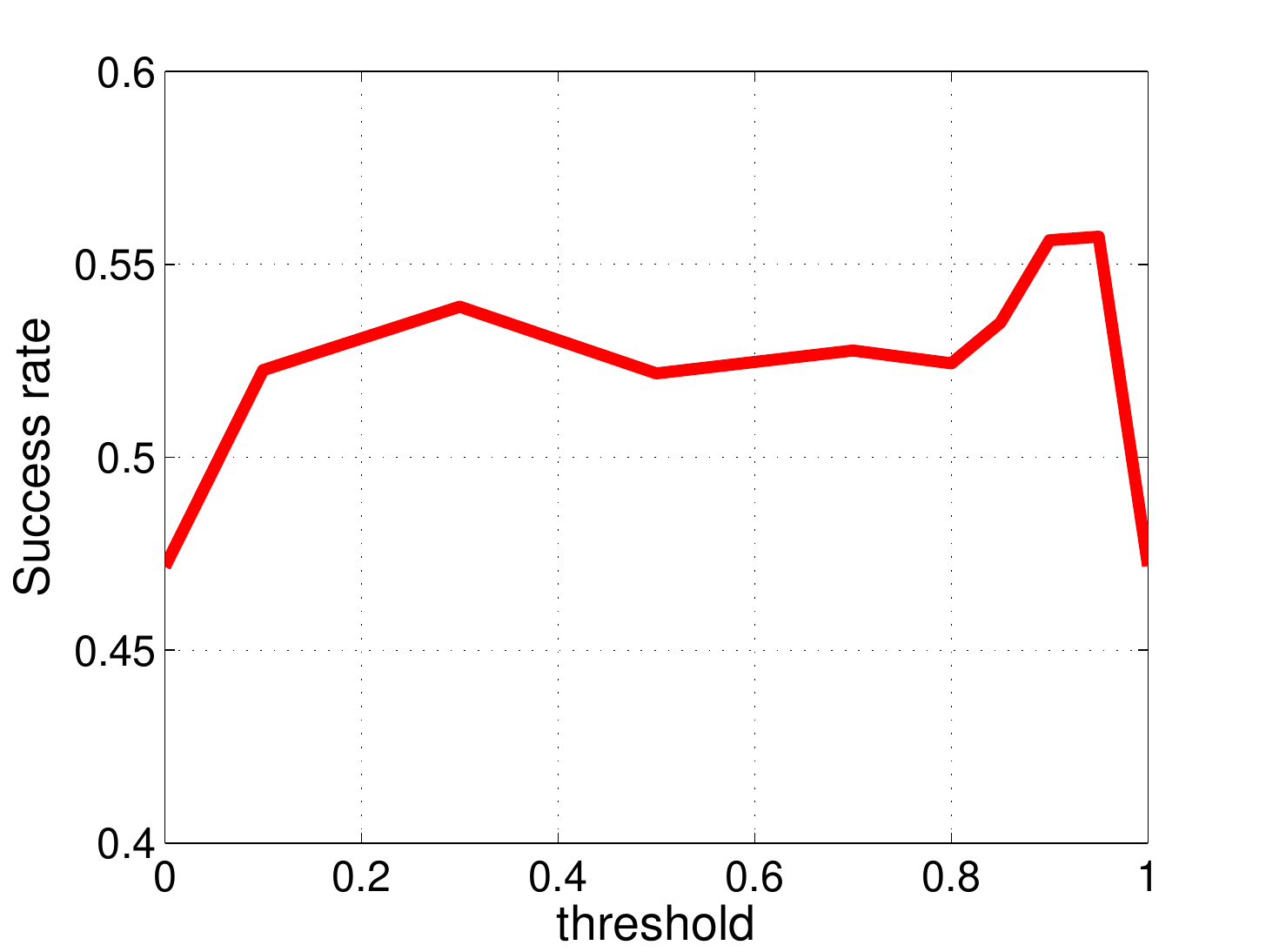}}
	\subfigure[Precision@20 for central pixel error curve]{\includegraphics[width=0.48\linewidth]{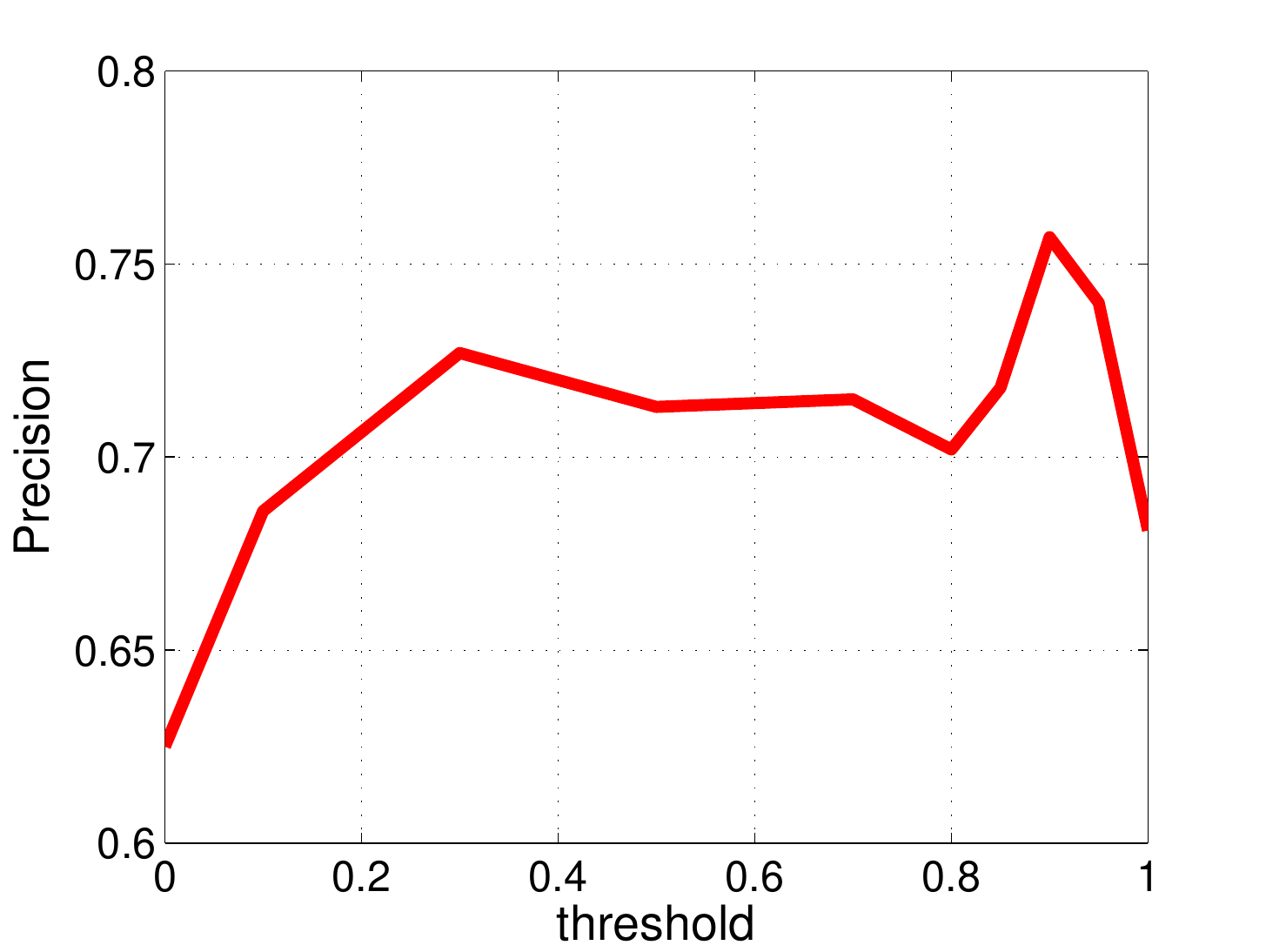}}
	\caption{ Results of varying the threshold for the first model update method.}
	\label{fig:updater_classificationScore}
\end{figure}

\begin{figure}[htb]
	\centering\vspace{-0.1cm}
	\subfigure[AUC of overlap rate]{\includegraphics[width=0.48\linewidth]{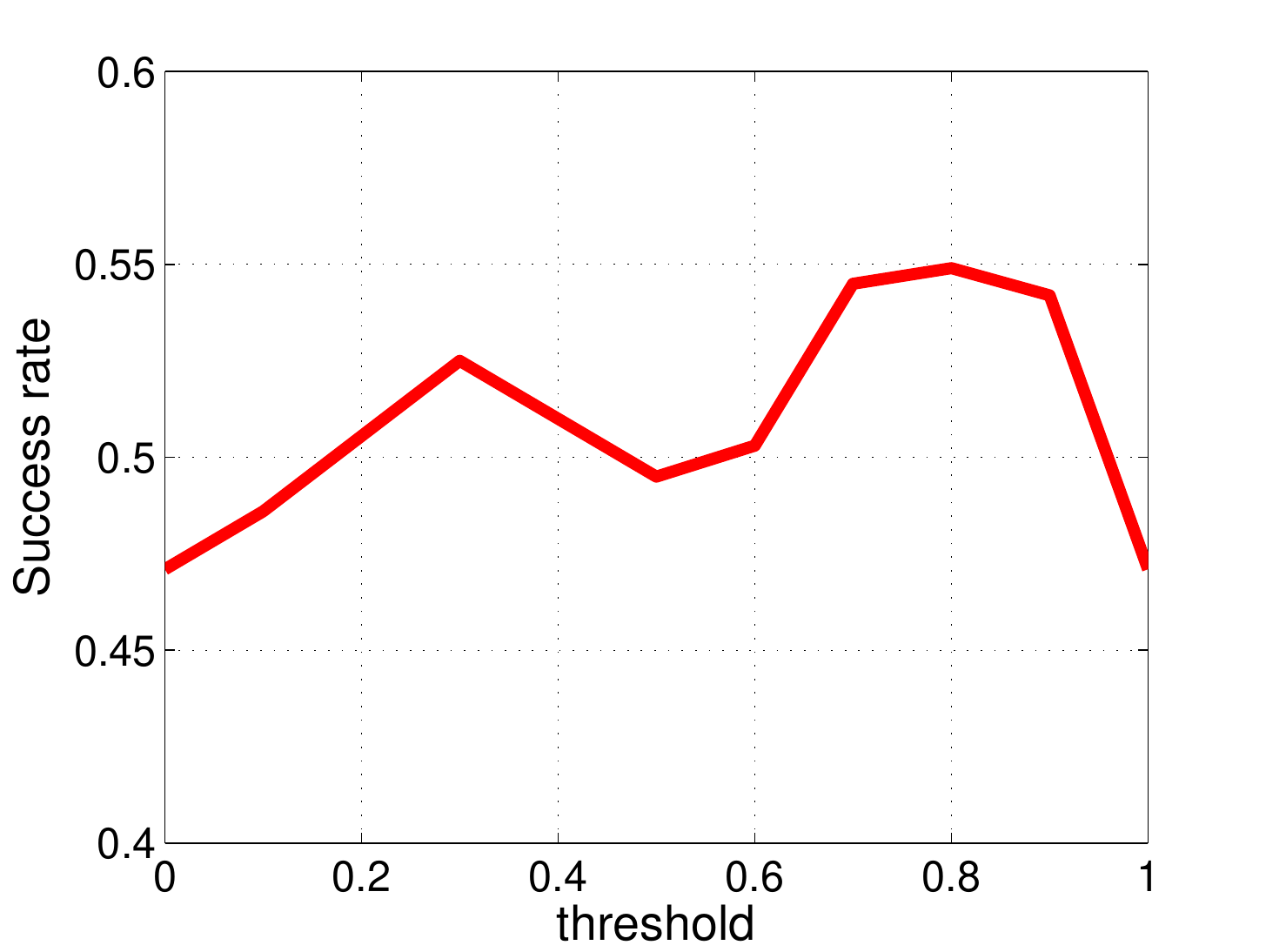}}
	\subfigure[Precision@20 for central pixel error curve]{\includegraphics[width=0.48\linewidth]{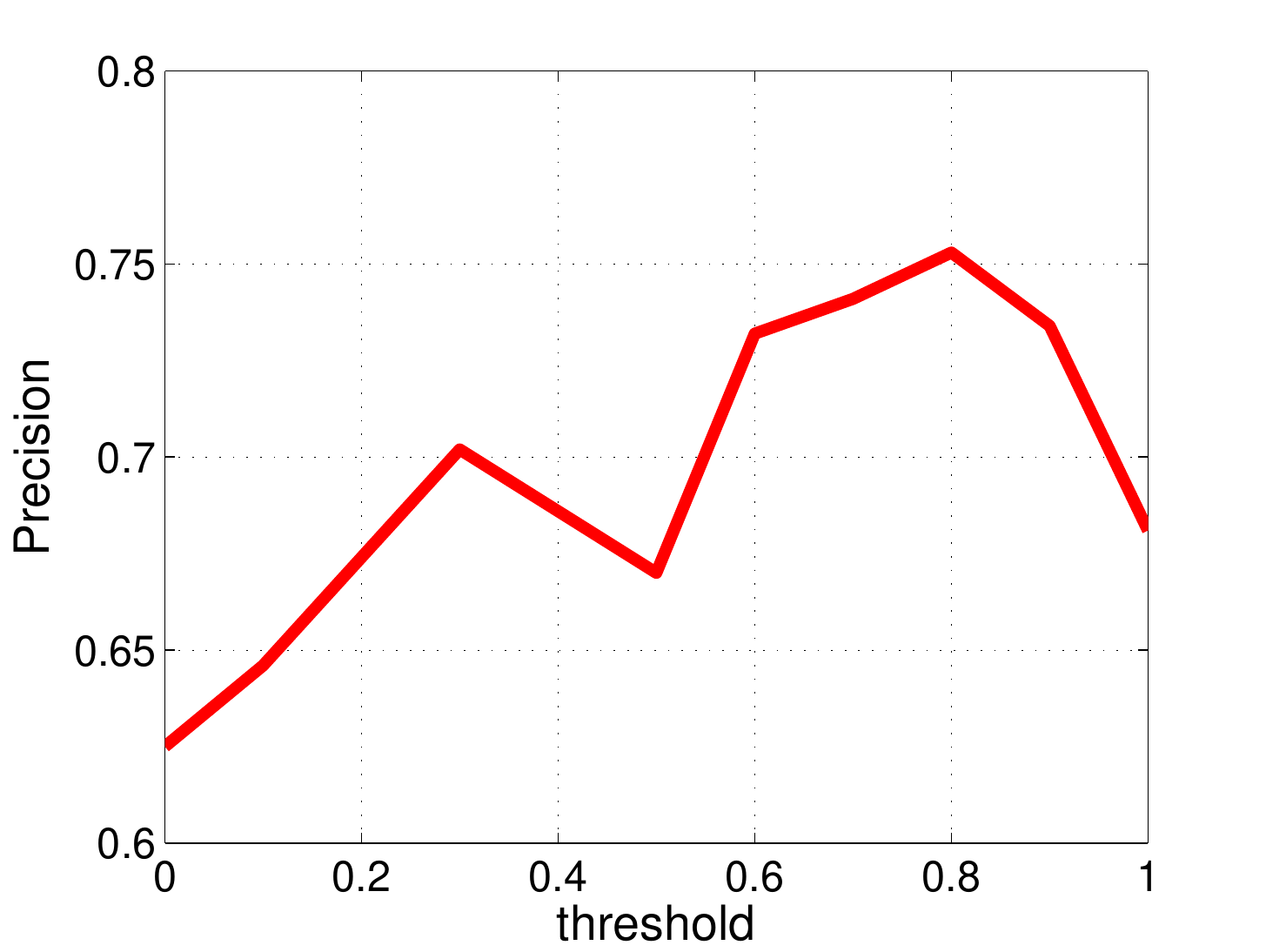}}
	\caption{ Results of varying the threshold for the second model update method.}\vspace{-0.1cm}
	\label{fig:updater_scoreDifference}
\end{figure}

We show the results of these two methods in Fig.~\ref{fig:updater_classificationScore} and Fig.~\ref{fig:updater_scoreDifference}.

Varying the threshold can indeed affect the results by more than $10\%$.  The best results for both methods are very similar, although the second method seems to give satisfactory results over a broader range of parameters.

Most research effort in this area focuses on generative trackers.  In~\cite{template}, Matthews \etal first empirically compared the effect of different template update strategies.  Following this work, Ross \etal proposed to use incremental PCA~\cite{ivt} for template update, Wang \etal showed the importance of sparsity and robustness~\cite{onndl} for this problem, and Xing \etal proposed to maintain three dictionaries of different lifespans~\cite{lifespan}.
However, the model updater is less studied in discriminative trackers. To the best of our knowledge, the only principled method for model updater is the one by~\cite{meem}. They proposed to use entropy minimization to identify reliable model update and discard the incorrect ones.
  
\paragraph{Our Findings:} \emph{Although implementation of the model updater is often treated as engineering tricks in papers especially for discriminative trackers, their impact on performance is usually very significant and hence is worth studying.  Unfortunately, very few work focuses on this component.}

\subsection{Ensemble Post-processor}
\begin{figure}[htb]
	\centering
	\includegraphics[width=0.495\linewidth]{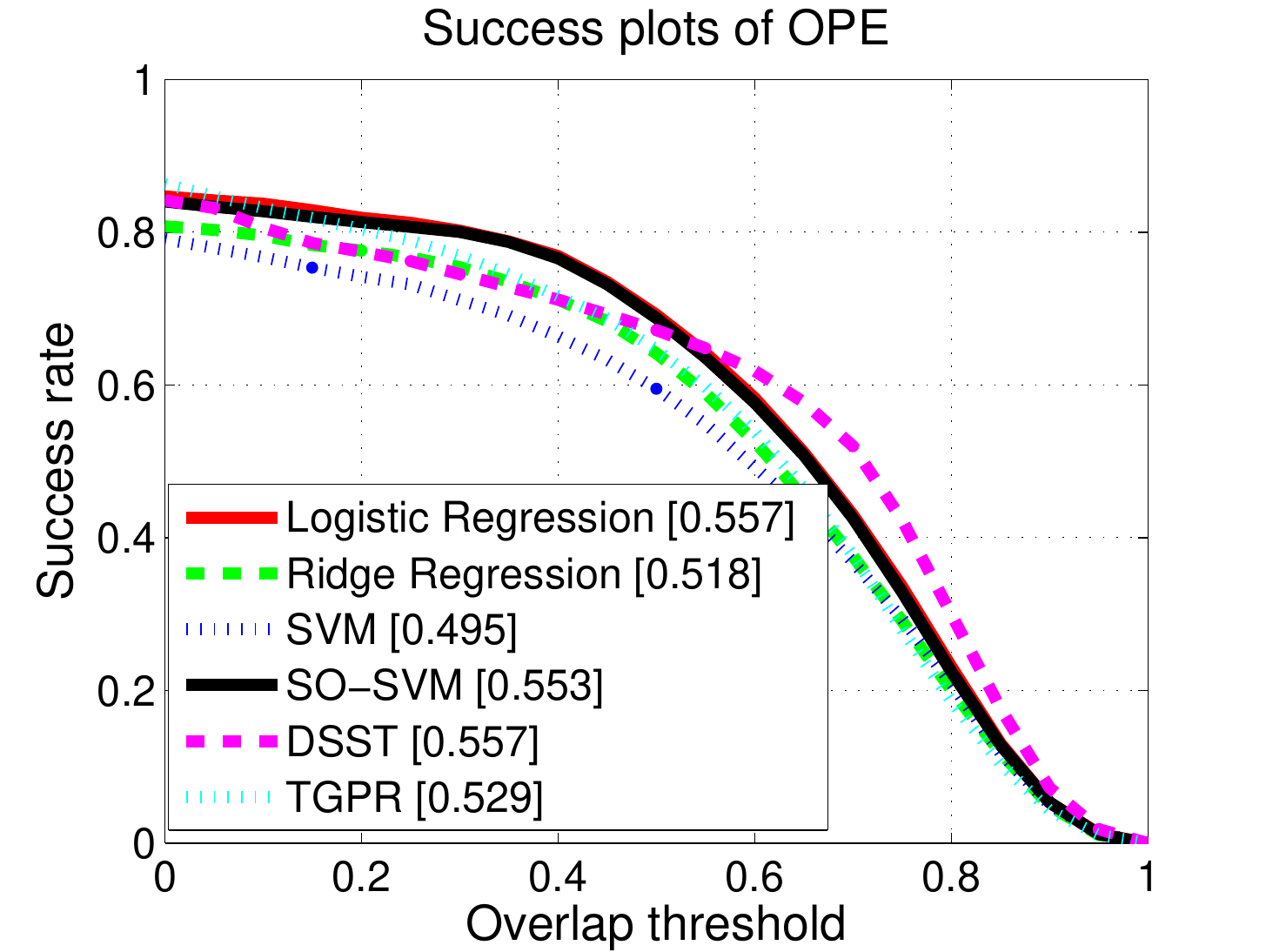}
	\includegraphics[width=0.495\linewidth]{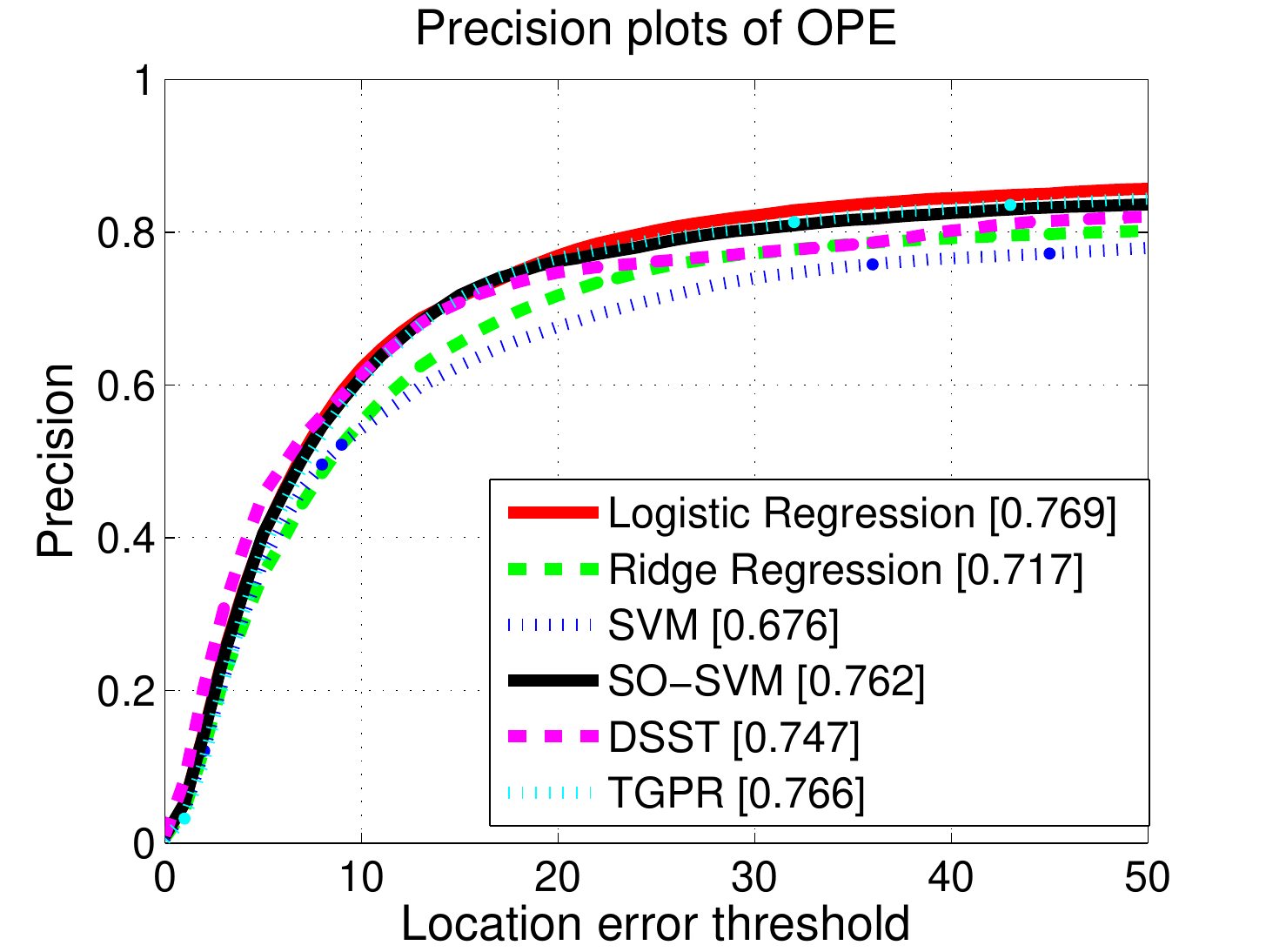}
	\caption{Results of individual trackers used in ensemble.}
	\label{fig:ensemble_individual}
\end{figure}

\begin{figure}[htb]
	\centering
	\includegraphics[width=0.495\linewidth]{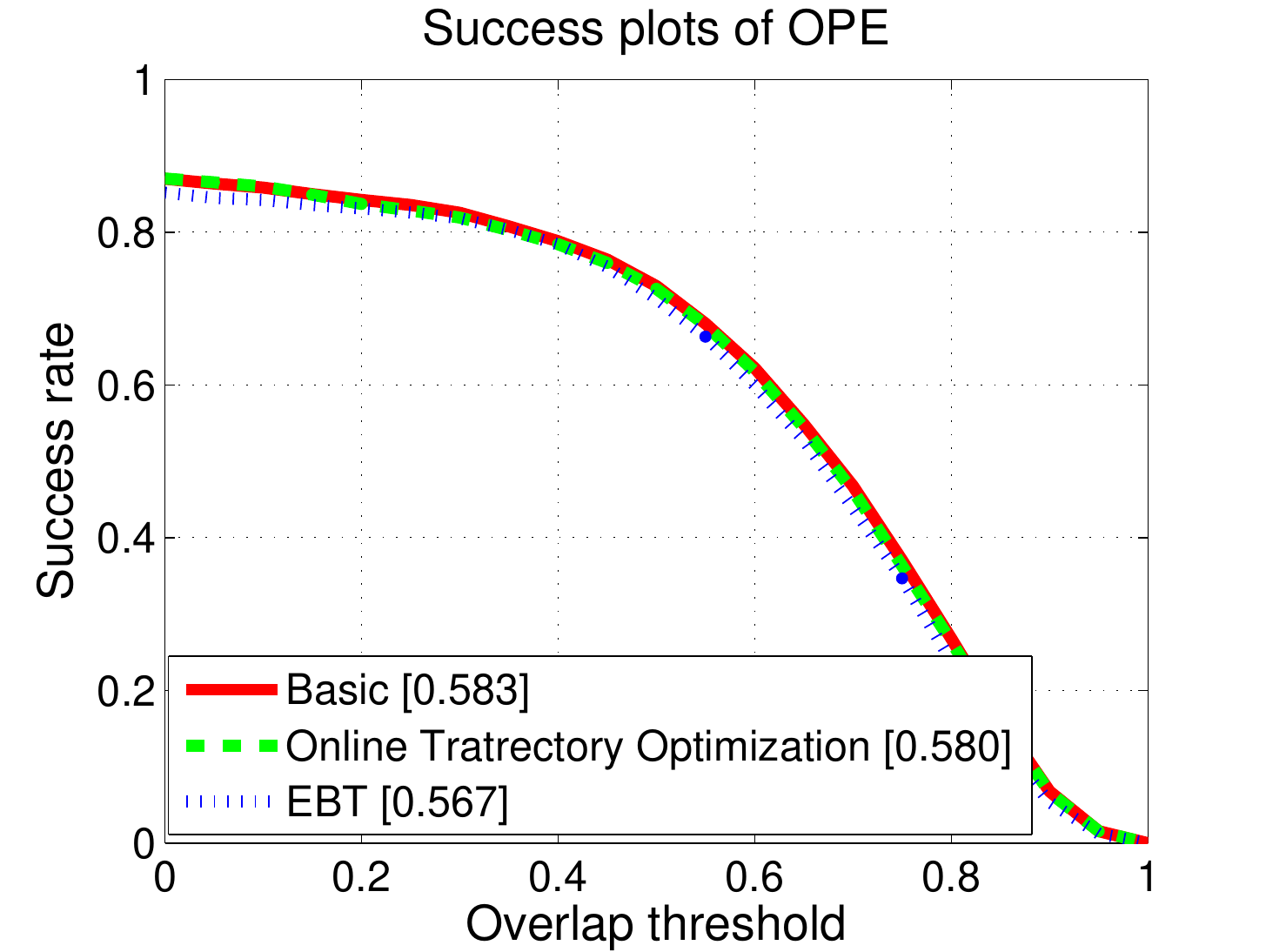}
	\includegraphics[width=0.495\linewidth]{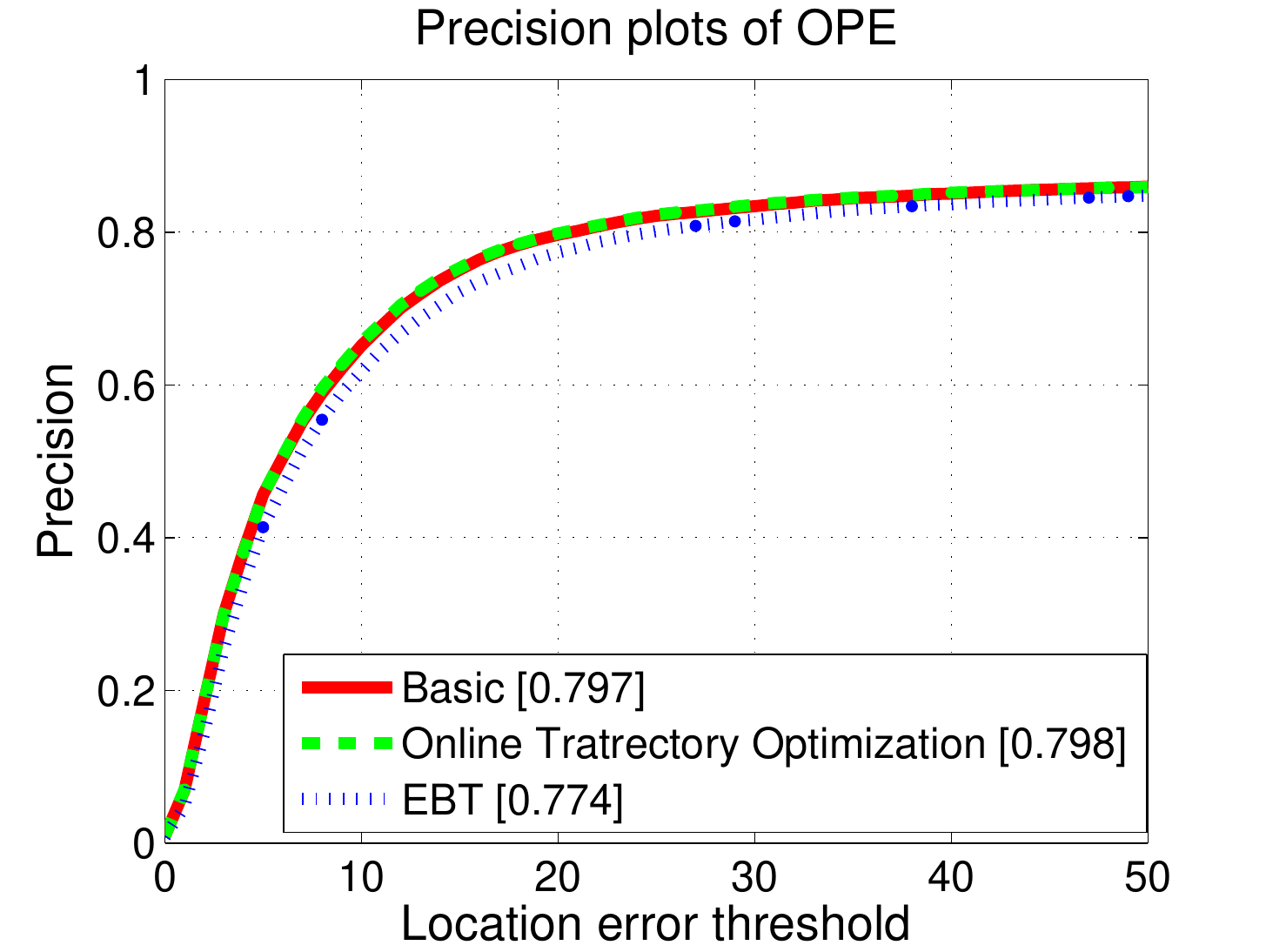}
	\caption{Results of ensemble when the individual trackers are of low diversity (the four different observation models from our framework). \emph{Basic} and \emph{Online Trajectory Optimization} methods are from~\cite{superior} and \emph{EBT} is from~\cite{ebt}.}
	\label{fig:ensemble_low_diversity}
\end{figure}

\begin{figure}[htb]
	\centering
	\includegraphics[width=0.495\linewidth]{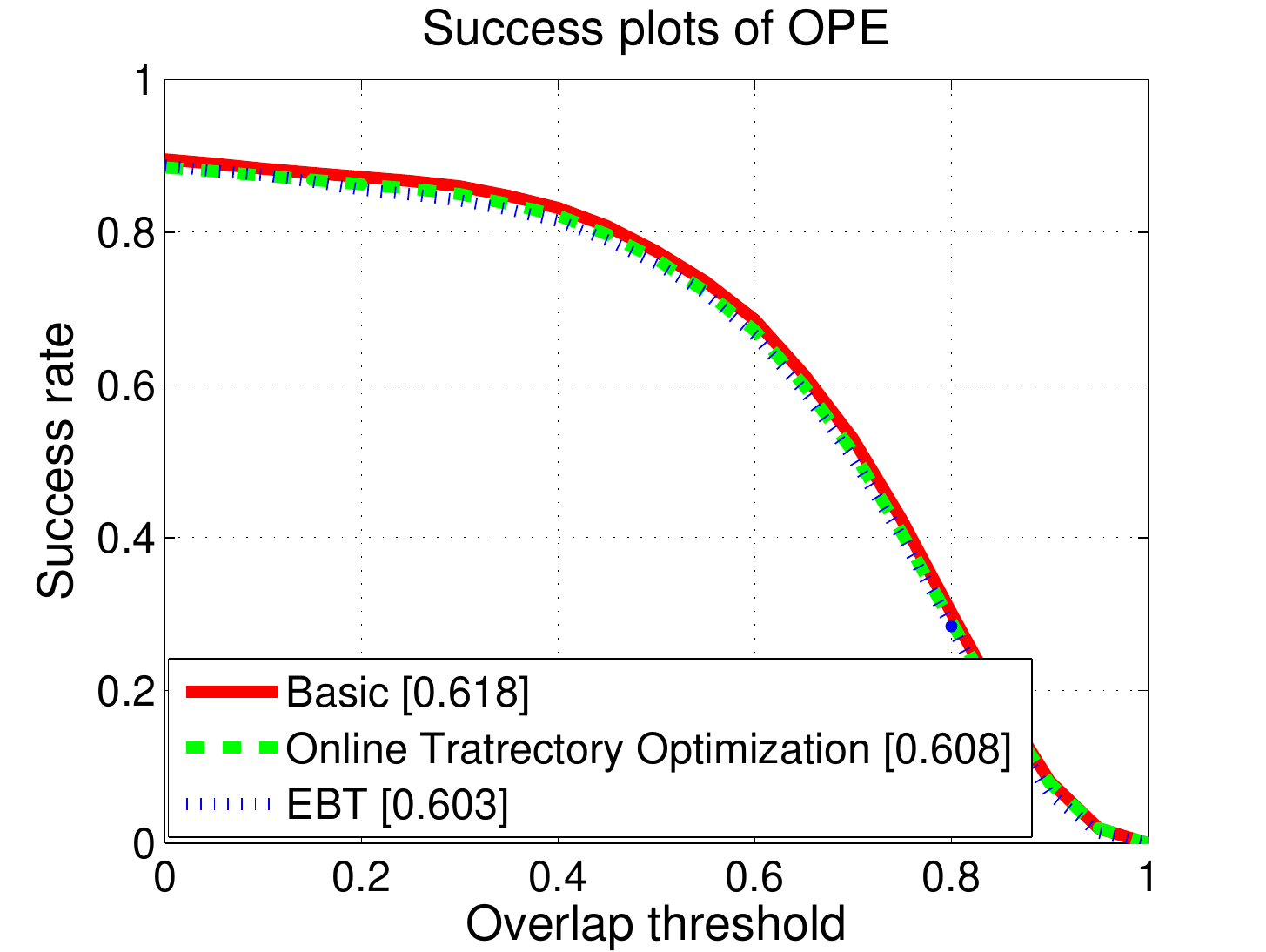}
	\includegraphics[width=0.495\linewidth]{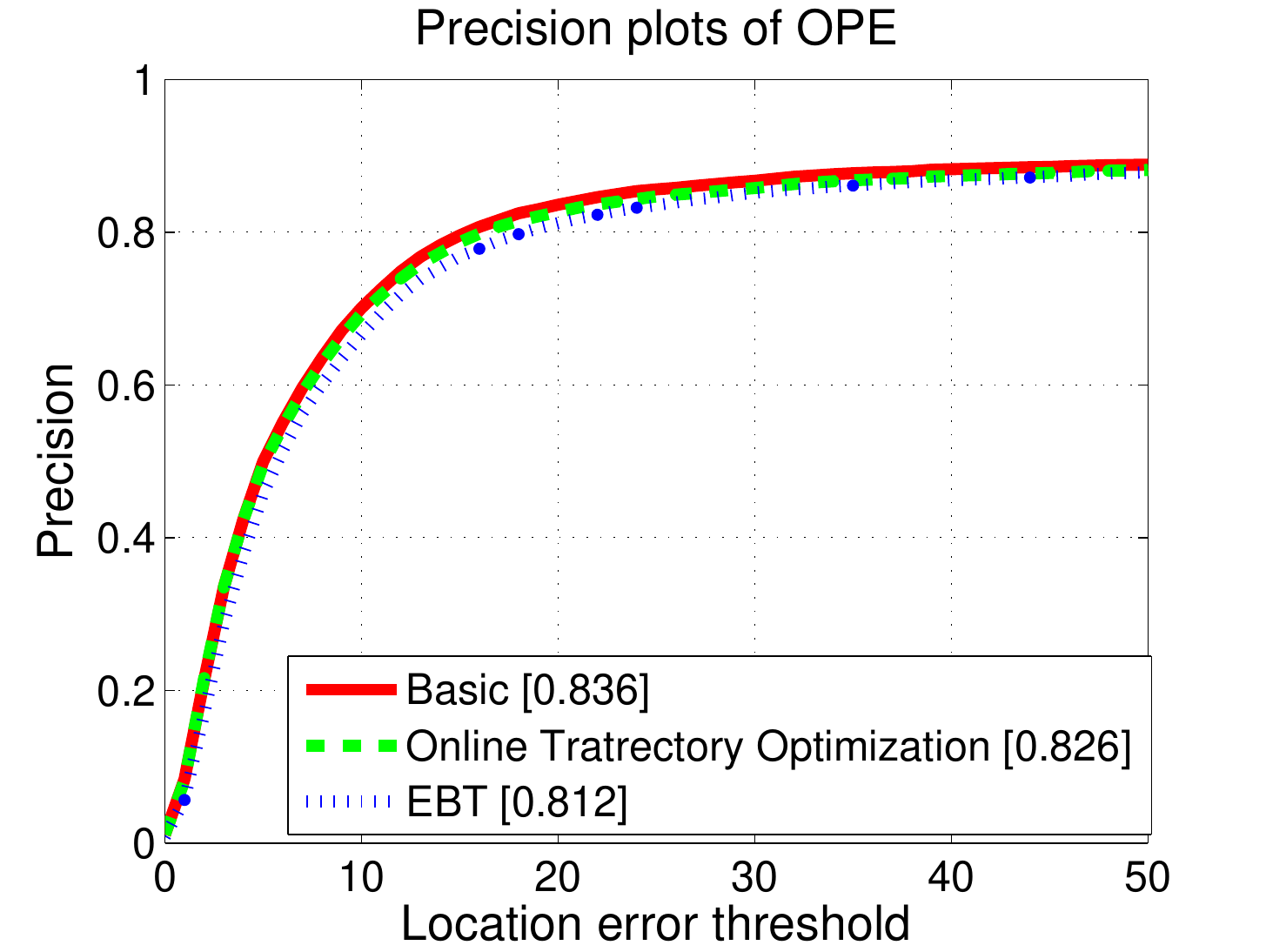}
	\caption{Results of ensemble when the individual trackers are of high diversity (all the six trackers). \emph{Basic} and \emph{Online Trajectory Optimization} methods are from~\cite{superior} and \emph{EBT} is from~\cite{ebt}.}
	\label{fig:ensemble_high_diversity}
\end{figure}

From the analysis above, we can see that the result of a single tracker can sometimes be very unstable in that the performance can vary a lot even under small perturbation of the parameters.  The purpose of taking the ensemble approach is to overcome this limitation.  We regard the ensemble as a post-processing component which treats the constituent trackers as blackboxes and takes only the bounding boxes returned by them as input.  This rationale is quite different from ensemble tracking~\cite{oab, semiB} which uses boosting to build a better observation model.  Our ensemble includes six trackers, with four of them corresponding to four different observation models in our framework and the other two are DSST~\cite{dsst} and TGPR~\cite{gpr}. We choose these two trackers because they are among the best-performing trackers, and their techniques are complementary to ours. We show the performance of individual trackers in Fig.~\ref{fig:ensemble_individual}. Their results are very competitive. For the ensemble, we consider two recent methods:
\begin{enumerate}
	\item The first one is from~\cite{superior}. This paper first proposed a loss function for bounding box majority voting and then extended it to incorporate tracker weights, trajectory continuity and removal of bad trackers. We adopt two methods from the paper: the basic model and online trajectory optimization.
	\item The second one is from~\cite{ebt}. The authors formulated the ensemble learning problem as a structured crowdsourcing problem which treats the reliability of each tracker as a hidden variable to be inferred. Then they proposed a factorial hidden Markov model that considers the temporal smoothness between frames. We adopt the basic model called ensemble based tracking (EBT) without self-correction.
\end{enumerate}
Since the four trackers from our framework are all using the same features and motion model, their diversity is somewhat limited. A main reason of including the last two trackers into the ensemble is to increase the diversity of the trackers, because diversity often plays an important role in increasing the effectiveness of an ensemble.  To investigate how diversity can affect the ensemble performance, we report two sets of results: with and without DSST and TGPR. Their results are shown in Fig.~\ref{fig:ensemble_low_diversity} and Fig.~\ref{fig:ensemble_high_diversity}, respectively.

We can see that diversity in the ensemble helps to achieve good results.  Both ensemble methods can significantly improve the results when the trackers have high diversity.  Even when the diversity is low, the ensemble does not impair the performance but still slightly outperforms the best single tracker.

\paragraph{Our Findings:} \emph{The ensemble post-processor can improve the performance substantially especially when the trackers have high diversity. This component is universal and effective yet it is least explored.}

\section {Limitations of Current Framework}
The primary goal of this work is to gain a deeper understanding into the different components of a visual tracking system, rather than trying to include all existing trackers into our framework.  Thus, inevitably, some excellent trackers are not represented in the current framework.  We list and discuss some of them here.

First, in some methods, several components are tightly coupled.  For example, in the classical mean-shift tracker~\cite{meanshift}, the observation model must be paired with a probabilistic map as output; in some part-based methods, such as~\cite{frag, asla}, the observation model must be designed in such a way to take the part information into consideration; and in the latest deep learning trackers~\cite{dlt, sodlt}, the feature extractor and observation model are combined into a unified deep learning framework for end-to-end learning.

Second, while accuracy is an important factor in visual tracking systems, it is certainly not the only one.  Speed is another important factor to consider in practice.  Since our framework is designed to be as universal and generic as possible to accommodate more, though not all, algorithms, we have not put much effort on optimizing the speed on purpose. Our best combination runs about 10fps in MATLAB. There exist some recent attempts that focus on developing fast tracking models.  For example, fast Fourier transform (FFT)~\cite{mosse} and circular matrices~\cite{kcf, dsst} are used to accelerate dense (kernelized) ridge regression. In their work, the motion model and observation model are coupled.  Although we could approximate their methods in our framework using sliding windows and ridge regression, such implementation would be much slower than that in the original paper.

\section {Conclusion and Future Work}

\epigraph{{\normalsize``God is in the details."}}{--- \textup{Ludwig Mies van der Rohe}}

In this paper, we have analyzed and identified some important factors for a good visual tracking system.  We show that if we design each component carefully, even some very elementary building blocks from textbooks can result in a tracker that is as competitive as state-of-the-art trackers.  By breaking a visual tracking system down into its constituent parts and analyzing each of them carefully, we have arrived at some interesting conclusions.  First, the feature extractor is the most important part of a tracker.  Second, the observation model is not that important if the features are good enough.  Third, the model updater can affect the result significantly, but currently there are not many principled ways for realizing this component.  Lastly, the ensemble post-processor is quite universal and effective.  Besides, we demonstrate that paying attention to some details of the motion model and model updater can significantly improve the performance.

Our work enlightens several interesting directions to pursue, including the development of lightweight and effective feature representations, principled ways of model update, and advanced ensemble methods.  It is our hope that, besides the observation model which has been the focus of many studies, other equally important components in tracking systems will attract more research attention as a consequence of our findings.

{\small 
\bibliographystyle{ieee}
\bibliography{tracking}

\begin{thebibliography}{10}\itemsep=-1pt

\bibitem{frag}
A.~Adam, E.~Rivlin, and I.~Shimshoni.
\newblock Robust fragments-based tracking using the integral histogram.
\newblock In {\em IEEE Conference on Computer Vision and Pattern Recognition},
  pages 798--805, 2006.

\bibitem{tutorial}
M.~Arulampalam, S.~Maskell, N.~Gordon, and T.~Clapp.
\newblock {A tutorial on particle filters for online nonlinear/non-Gaussian
  Bayesian tracking}.
\newblock {\em IEEE Transactions on Signal Processing}, 50(2):174--188, 2002.

\bibitem{mil}
B.~Babenko, M.~Yang, and S.~Belongie.
\newblock Robust object tracking with online multiple instance learning.
\newblock {\em IEEE Transactions on Pattern Analysis and Machine Intelligence},
  33(8):1619--1632, 2011.

\bibitem{superior}
C.~Bailer, A.~Pagani, and D.~Stricker.
\newblock A superior tracking approach: Building a strong tracker through
  fusion.
\newblock In {\em European Conference on Computer Vision}, pages 170--185.
  2014.

\bibitem{mosse}
D.~S. Bolme, J.~R. Beveridge, B.~A. Draper, and Y.~M. Lui.
\newblock Visual object tracking using adaptive correlation filters.
\newblock In {\em IEEE Conference on Computer Vision and Pattern Recognition},
  pages 2544--2550, 2010.

\bibitem{metric}
L.~{\v{C}}ehovin, A.~Leonardis, and M.~Kristan.
\newblock Visual object tracking performance measures revisited.
\newblock {\em arXiv preprint arXiv:1502.05803}, 2015.

\bibitem{meanshift}
D.~Comaniciu, V.~Ramesh, and P.~Meer.
\newblock Real-time tracking of non-rigid objects using mean shift.
\newblock In {\em IEEE Conference on Computer Vision and Pattern Recognition},
  pages 142--149, 2000.

\bibitem{hog}
N.~Dalal and B.~Triggs.
\newblock Histograms of oriented gradients for human detection.
\newblock In {\em IEEE Conference on Computer Vision and Pattern Recognition},
  pages 886--893, 2005.

\bibitem{dsst}
M.~Danelljan, G.~H{\"a}ger, F.~S. Khan, and M.~Felsberg.
\newblock Accurate scale estimation for robust visual tracking.
\newblock In {\em European Conference on Computer Vision}, 2014.

\bibitem{adaptiveColor}
M.~Danelljan, F.~S. Khan, M.~Felsberg, and J.~v.~d. Weijer.
\newblock Adaptive color attributes for real-time visual tracking.
\newblock In {\em IEEE Conference on Computer Vision and Pattern Recognition},
  pages 1090--1097, 2014.

\bibitem{gpr}
J.~Gao, H.~Ling, W.~Hu, and J.~Xing.
\newblock Transfer learning based visual tracking with {Gaussian} processes
  regression.
\newblock In {\em European Conference on Computer Vision}, pages 188--203.
  2014.

\bibitem{oab}
H.~Grabner, M.~Grabner, and H.~Bischof.
\newblock Real-time tracking via on-line boosting.
\newblock In {\em British Machine Vision Conference}, pages 47--56, 2006.

\bibitem{semiB}
H.~Grabner, C.~Leistner, and H.~Bischof.
\newblock Semi-supervised on-line boosting for robust tracking.
\newblock In {\em European Conference on Computer Vision}, pages 234--247,
  2008.

\bibitem{struck}
S.~Hare, A.~Saffari, and P.~H. Torr.
\newblock Struck: Structured output tracking with kernels.
\newblock In {\em Inernational Conference on Computer Vision}, pages 263--270,
  2011.

\bibitem{kcf}
J.~F. Henriques, R.~Caseiro, P.~Martins, and J.~Batista.
\newblock High-speed tracking with kernelized correlation filters.
\newblock {\em arXiv preprint arXiv:1404.7584}, 2014.

\bibitem{cnnt}
S.~Hong, T.~You, S.~Kwak, and B.~Han.
\newblock Online tracking by learning discriminative saliency map with
  convolutional neural network.
\newblock {\em arXiv preprint arXiv:1502.06796}, 2015.

\bibitem{asla}
X.~Jia, H.~Lu, and M.~Yang.
\newblock Visual tracking via adaptive structural local sparse appearance
  model.
\newblock In {\em IEEE Conference on Computer Vision and Pattern Recognition},
  pages 1822--1829, 2012.

\bibitem{vot14}
M.~Kristan and \etal.
\newblock The visual object tracking {VOT2014} challenge results.
\newblock In {\em European Conference on Computer Vision Workshop}, 2014.

\bibitem{nuspro}
A.~Li, M.~Lin, Y.~Wu, M.-H. Yang, and S.~Yan.
\newblock {NUS-PRO}: A new visual tracking challenge.
\newblock {\em To Appear in IEEE Transactions on Pattern Analysis and Machine
  Intelligence}, 2015.

\bibitem{klt1}
B.~D. Lucas and T.~Kanade.
\newblock An iterative image registration technique with an application to
  stereo vision.
\newblock In {\em International Joint Conference on Artificial Intelligence},
  pages 674--679, 1981.

\bibitem{onlineDL}
J.~Mairal, F.~Bach, J.~Ponce, and G.~Sapiro.
\newblock Online learning for matrix factorization and sparse coding.
\newblock {\em Journal of Machine Learning Research}, 11(1):19--60, 2010.

\bibitem{template}
I.~Matthews, T.~Ishikawa, and S.~Baker.
\newblock The template update problem.
\newblock {\em IEEE Transactions on Pattern Analysis and Machine Intelligence},
  26(6):810--815, 2004.

\bibitem{l1t}
X.~Mei and H.~Ling.
\newblock Robust visual tracking using {$l_1$} minimization.
\newblock In {\em Inernational Conference on Computer Vision}, pages
  1436--1443, 2009.

\bibitem{bias}
Y.~Pang and H.~Ling.
\newblock Finding the best from the second bests-inhibiting subjective bias in
  evaluation of visual tracking algorithms.
\newblock In {\em Inernational Conference on Computer Vision}, pages
  2784--2791, 2013.

\bibitem{defenseColor}
H.~Possegger, T.~Mauthner, and H.~Bischof.
\newblock In defense of color-based model-free tracking.
\newblock In {\em IEEE Conference on Computer Vision and Pattern Recognition},
  2015.

\bibitem{ivt}
D.~Ross, J.~Lim, R.~Lin, and M.~Yang.
\newblock Incremental learning for robust visual tracking.
\newblock {\em International Journal of Computer Vision}, 77(1):125--141, 2008.

\bibitem{survey}
A.~Smeulders, D.~Chu, R.~Cucchiara, S.~Calderara, A.~Dehghan, and M.~Shah.
\newblock Visual tracking: An experimental survey.
\newblock {\em IEEE Transactions on Pattern Analysis and Machine Intelligence},
  36(7), 2014.

\bibitem{rgbd}
S.~Song and J.~Xiao.
\newblock Tracking revisited using {RGBD} camera: Baseline and benchmark.
\newblock In {\em Inernational Conference on Computer Vision}, pages 233--240,
  2013.

\bibitem{spllt}
J.~Supancic and D.~Ramanan.
\newblock Self-paced learning for long-term tracking.
\newblock In {\em IEEE Conference on Computer Vision and Pattern Recognition},
  pages 2379--2386, 2013.

\bibitem{klt2}
C.~Tomasi and T.~Kanade.
\newblock Detection and tracking of point features.
\newblock Technical Report CMU-CS-91-132, School of Computer Science, Carnegie
  Mellon Univ. Pittsburgh, 1991.

\bibitem{haar}
P.~Viola and M.~Jones.
\newblock Rapid object detection using a boosted cascade of simple features.
\newblock In {\em IEEE Conference on Computer Vision and Pattern Recognition},
  pages 511--518, 2001.

\bibitem{lsst}
D.~Wang, H.~Lu, and M.-H. Yang.
\newblock Least soft-threshold squares tracking.
\newblock In {\em IEEE Conference on Computer Vision and Pattern Recognition},
  pages 2371--2378, 2013.

\bibitem{sodlt}
N.~Wang, S.~Li, A.~Gupta, and D.-Y. Yeung.
\newblock Transferring rich feature hierarchies for robust visual tracking.
\newblock {\em arXiv preprint arXiv:1501.04587}, 2015.

\bibitem{onndl}
N.~Wang, J.~Wang, and D.-Y. Yeung.
\newblock Online robust non-negative dictionary learning for visual tracking.
\newblock In {\em Inernational Conference on Computer Vision}, pages 657--664,
  2013.

\bibitem{dlt}
N.~Wang and D.-Y. Yeung.
\newblock Learning a deep compact image representation for visual tracking.
\newblock In {\em The Conference on Neural Information Processing Systems},
  pages 809--817, 2013.

\bibitem{ebt}
N.~Wang and D.-Y. Yeung.
\newblock Ensemble-based tracking: Aggregating crowdsourced structured time
  series data.
\newblock In {\em ICML}, pages 1107--1115, 2014.

\bibitem{onlinesvm}
Z.~Wang and S.~Vucetic.
\newblock Online training on a budget of support vector machines using twin
  prototypes.
\newblock {\em Statistical Analysis and Data Mining: The ASA Data Science
  Journal}, 3(3):149--169, 2010.

\bibitem{benchmark}
Y.~Wu, J.~Lim, and M.-H. Yang.
\newblock Online object tracking: A benchmark.
\newblock In {\em IEEE Conference on Computer Vision and Pattern Recognition},
  2013.

\bibitem{benchmark2}
Y.~Wu, J.~Lim, and M.-H. Yang.
\newblock Object tracking benchmark.
\newblock {\em To Appear in IEEE Transactions on Pattern Analysis and Machine
  Intelligence}, 2015.

\bibitem{lifespan}
J.~Xing, J.~Gao, B.~Li, W.~Hu, and S.~Yan.
\newblock Robust object tracking with online multi-lifespan dictionary
  learning.
\newblock In {\em Inernational Conference on Computer Vision}, pages 665--672,
  2013.

\bibitem{meem}
J.~Zhang, S.~Ma, and S.~Sclaroff.
\newblock {MEEM}: Robust tracking via multiple experts using entropy
  minimization.
\newblock In {\em European Conference on Computer Vision}, pages 188--203.
  2014.

\bibitem{scm}
W.~Zhong, H.~Lu, and M.-H. Yang.
\newblock Robust object tracking via sparsity-based collaborative model.
\newblock In {\em IEEE Conference on Computer Vision and Pattern Recognition},
  pages 1838--1845, 2012.

\end{thebibliography}
}

\end{document}